\documentclass[a4paper,fleqn]{cas-sc}

\usepackage[authoryear]{natbib}
\usepackage{apalike}
\usepackage{amsmath}
\usepackage{fancyhdr}
\usepackage{threeparttable}
\usepackage{multirow}
\usepackage{pdflscape}
\usepackage{array}
\usepackage{amsmath}
\usepackage{float}
\usepackage{rotating}
\usepackage[section]{placeins}

\begin{document}
\let\WriteBookmarks\relax
\def\floatpagepagefraction{1}
\def\textpagefraction{.001}

\shortauthors{J. Long, et al.}
\shorttitle{Large-Scale, Pixel-Wise Crop Mapping}

\title [mode = title]{From Time-series Generation, Model Selection to Transfer Learning: A Comparative Review of Pixel-wise Approaches for Large-scale Crop Mapping}

\author[1]{Judy Long}
\credit{Conceptualization, Methodology, Software, Writing - Original Draft Preparation.}

\author[2]{Tao Liu}[
      orcid=0000-0001-5917-8303]
\credit{Implementation Support, Interpretation, Writing - Review and Editing.}
\cormark[1]
\cortext[cor1]{Corresponding author}
\ead{taoliu@uga.edu}
\ead[url]{https://warnell.uga.edu/directory/people/tao-liu}

\author[3]{Sean Alexander Woznicki}
\credit{Writing - Review and Editing.}

\author[4]{Miljana Marković}
\credit{Results Validation, Writing - Review and Editing.}

\author[4]{Oskar Marko}
\credit{Writing - Review and Editing.}

\author[5]{Molly Sears}
\credit{Writing - Review and Editing.}

\affiliation[1]{organization={College of Forest Resources and Environmental Science, Michigan Technological University},
    addressline={1400 Townsend Drive}, 
    city={Houghton},
    postcode={MI 49931}, 
    country={United States}}

\affiliation[2]{organization={Warnell School of Forestry and Natural Resources, University of Georgia},
    addressline={180 East Green Street}, 
    city={Athens},
    postcode={GA 30602}, 
    country={United States}}

\affiliation[3]{organization={Annis Water Resources Institute, Grand Valley State University},
    addressline={740 West Shoreline Drive}, 
    city={Muskegon},
    postcode={MI 49441}, 
    country={United States}}

\affiliation[4]{organization={BioSense Institute},
    addressline={Dr Zorana Đinđića 1}, 
    city={Novi Sad},
    postcode={21000}, 
    country={Serbia}}

\affiliation[5]{organization={Department of Agricultural, Food, and Resource Economics, Michigan State University},
    addressline={426 Auditorium Road}, 
    city={East Lansing},
    postcode={48824}, 
    country={United States}}

\begin{highlights}

\item This study reviewed current practices in large-scale, pixel-wise crop mapping and transfer learning workflows.

\item Evaluated six preprocessing methods, eleven model architectures, and three transfer learning strategies through systematic experiments. 

\item Fine-scale (7-day) interval preprocessing paired with Transformer models consistently delivered optimal performance for both supervised and transferable workflows.

\item Transfer learning techniques substantially enhanced workflow adaptability, with unsupervised domain adaptation (UDA) being effective for homogeneous crop classes while fine-tuning remains robust across diverse scenarios.

\item Transfer learning that matches the level of domain shift is a viable alternative to achieve crop mapping when the available sample is below a certain threshold, and supervised classification becomes unstable.

\end{highlights}


\begin{abstract}
Crop mapping involves identifying and classifying crop types using spatial data, primarily derived from remote sensing imagery. Recent advances in machine learning have improved crop mapping efficiency, allowing scalable, automated mapping using ground truth data or prior knowledge. This study presents the first comprehensive review of large-scale, pixel-wise crop mapping workflows, encompassing both conventional supervised methods and emerging transfer learning approaches. 

To identify the optimal time-series generation approaches and supervised crop mapping models, we conducted systematic experiments, compared six widely adopted satellite image-based preprocessing methods, including raw imagery, two Whittaker-Eilers smoothers, weekly and monthly linear resampling, and phenological peak extraction, alongside eleven supervised classification models, including Random Forest (RF), Recurrent Neural Network (RNN) variants, attention-based RNN variants, and Transformers. Additionally, we assessed the synergistic impact of varied training sample sizes and combinations of spectral, vegetation and water indices (VIs), and synthetic aperture radar (SAR) variables. Moreover, we identified optimal transfer learning techniques for different magnitudes of domain shift. Direct transfer workflows were evaluated across spatial, temporal, and sensor domains. Fine-tuning and an unsupervised domain adaptation (UDA) method, Domain-Adversarial Neural Networks (DANN), were assessed under varying spatial shift conditions. The evaluation of optimal methods was conducted across five diverse agricultural sites (approximately 185 km by 170 km each) in the U.S. Soybean Belt and Vojvodina, Serbia. Landsat 8 served as the primary satellite data source; Landsat 7 and Sentinel-1 supported transfer and fusion evaluations separately. Labels of corn, soybean, and other, come from Cropland Data Layer (CDL) trusted pixels and field surveys. 

Our findings reveal three key insights. First, fine-scale (7-day) interval preprocessing paired with Transformer models consistently delivered optimal performance for both supervised and transferable workflows. RF offered rapid training and competitive performance in conventional supervised learning and direct transfer to similar domains. Second, transfer learning techniques substantially enhanced workflow adaptability, with UDA being effective for homogeneous crop classes while fine-tuning remains robust across diverse scenarios, including highly divergent domains. Finally, workflow choice depends heavily on the availability of labeled samples. With a sufficient sample size (e.g., approximately 3,000 samples for each class within a typical Landsat 8 scene area in the Corn-Soybean Belt), supervised training typically delivers more accurate and generalizable results. Below a certain threshold (which varies by region and crop complexity), supervised classification becomes unstable. Transfer learning that matches the level of domain shift is a viable alternative to achieve crop mapping. This review synthesizes current good practices and provides guidance for developing transferable pixel-wise large-scale crop mapping workflows. All code is publicly available to encourage reproducibility practice.

\end{abstract}


\begin{keywords}
Crop mapping \sep Pixel-wise \sep Large-scale \sep Deep learning \sep Transfer learning \sep Time series
\end{keywords}

\maketitle

\section{Introduction}

\subsection{Recent advances in remote sensing for crop mapping}

Crop mapping involves identifying and classifying crop types using spatial data, primarily derived from remote sensing imagery. Initially, cropland surveys relied on manual field inspections, which were challenging to standardize and scale. Since the 1970s, Earth observation missions such as Landsat (1972), SPOT (1986), MODIS (1999), Gaofen (2013), and Sentinel (2014) have provided extensive global remote sensing datasets, enabling large-scale environmental monitoring \citep{LentonTimothyM24, ChenJinyue22, ThiHuyen22}, agricultural yield estimation \citep{KhakiSaeed21}, and disaster management \citep{DiwatePranaya25, HakkenbergChristopherR24}. The introduction of high-resolution satellite imagery facilitated the integration of ground surveys with expert interpretation \citep{LillesandThomas15}, establishing workflows suitable for broad-scale cropland classification.

Recent advancements in machine learning (ML) and deep learning (DL) have enabled scalable and automated crop mapping, which has significantly enhanced crop monitoring efficiency. Advanced supervised learning combined with intensive field surveys supports timely annual or even in-season crop mapping \citep{KonduriVenkataShashank20}, notably in the U.S. \citep{BoryanClaire11} and Canada \citep{FisetteT13}, and accelerated data-driven decision-making and precision agriculture. Emerging transfer learning reduces the amount of needed crop type training data by transferring prior knowledge \citep{ArturN21}, thus has the potential to enable crop mapping in data-scarce regions. The availability of sparse, point- or parcel-based reference data led to the widespread adoption of pixel-wise and object-based classification, whereas patch-based methods (semantic segmentation) were developed in certain regions where complete field boundary information was available.

\begin{figure}
	\centering
		\includegraphics[scale=.52]{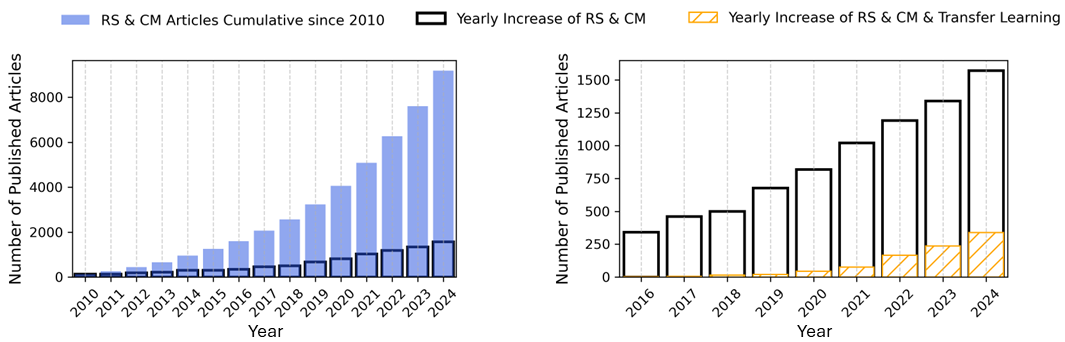}
	\caption{From 2010 to 2024, the number of studies integrating remote sensing with crop mapping increased by approximately 20\% annually. From 2016 onward, the use of transfer learning in this field grew at a geometric mean annual rate of 81\%. Data were retrieved from Google Scholar using two sets of search terms: (a) "Remote Sensing" AND "Crop Mapping" and (b) "Remote Sensing" AND "Crop Mapping" AND "Transfer Learning." Abbreviations: RS, remote sensing; CM, crop mapping.}
	\label{fig:1}
\end{figure}

Cloud-based platforms, particularly Google Earth Engine (GEE), have further advanced the accessibility and usability of large-scale geospatial archives covering over four decades \citep{GorelickNoel17}. These platforms facilitate seamless integration of remote sensing and ML workflows, removing traditional barriers such as data storage and computational limitations, and thereby fostering innovation \citep{YangLiping22}. Between 2010 and 2024, the integration of remote sensing with crop mapping research increased by approximately 20\% annually (Fig.~\ref{fig:1}); also, from 2016 onward, the use of transfer learning in this domain grew at a geometric mean annual rate of 81\%, these explosive increase driven by rapidly evolving ML and DL algorithms, widespread adoption of GEE, and increased availability of remote sensing products, including Landsat 8/9, Sentinel-2, and cost-effective airborne imagery.

However, inconsistencies in satellite image preprocessing methods, feature selection strategies, and model architectures have impeded the establishment of robust and scalable supervised crop mapping workflows. The lack of a comprehensive review of transfer learning applications also makes the optimal transfer learning workflow for crop mapping unclear. Herein, we present the first comprehensive review that covers both conventional supervised learning and emerging transfer learning approaches in large-scale, pixel-wise crop mapping. By reviewing existing preprocessing techniques (Section ~\ref{sec:intro_preprocessing}) and classification approaches (Section ~\ref{sec:intro_model}), we identified key research gaps and opportunities (Section ~\ref{sec:Intro_gaps}) that inform the focus of future studies.

\subsection{Satellite image preprocessing techniques} \label{sec:intro_preprocessing}

Time-series remote sensing data capture seasonal spectral changes, which offer advantages over single-date observations and are critical for distinguishing crop types. In this study, preprocessing refers to the reconstruction of time series from Level-2 surface reflectance products, involving noise removal and temporal standardization to produce consistent, cloud-free image sequences for model training. Table~\ref{tab:1} summarizes the preprocessing methods that have undergone continuous refinement in recent years. In addition to using single-source satellite imagery, some studies argued that adding complementary variables can further enhance data richness (Table~\ref{tab:2}).

\begin{table}[ht]
    \centering
    \caption{Time series reconstruction methods for crop mapping}
    \renewcommand{\arraystretch}{1.5}
    {\footnotesize
    \begin{tabular}{m{2cm} >{\centering\arraybackslash}m{1.5cm} >{\centering\arraybackslash}m{1.5cm} >{\arraybackslash}m{6cm} >{\centering\arraybackslash}m{3cm}}
        \hline
        \textbf{Method} & \textbf{Date Interval} & \textbf{Noise} & \textbf{Advantage} & \textbf{Example} \\
        \hline
        Raw data & Unequal & Retained & + Original spectral-temporal information. & \citep{Li24} \\
        Interpolation & Unequal & Removed & + Original acquisition date. & \citep{Cai18} \\
        Smoothing & Unequal & Removed & + Outliers are eliminated. & \citep{Khanal20, Feng23, Xuan23} \\
        Resampling & Equal & Removed & + Allows algorithms to scale to large areas. & \citep{PatrickG19, Yang24} \\
        \hline
    \end{tabular}%
    }
    \label{tab:1}
\end{table}

\begin{table}[ht]
    \centering
    \caption{Other variables integrated or augmented with the time series}
    {\footnotesize
    \begin{tabular}{m{4cm} m{8cm} >{\centering\arraybackslash}m{3cm}}
        \hline
        \textbf{Variable}  & \textbf{Advantage} & \textbf{Example} \\
        \hline
       Vegetation index & + Proxies for vegetation greenness and vigor.  &\citep{LihengZhong14, Xuan23} \\
        Statistical description  & + Summarizes central tendency, dispersion, and shape of data. & \citep{Xuan23} \\
         Deep feature representation & + Mining informative deep features. & \citep{XuYijia24, WenfangZ24} \\
         Multi-source variables & + Enhance spatial and temporal data richness. & \citep{YangC21} \\
         Environmental variables & + Indicate crop growth and planting decisions. & \citep{LihengZhong14}\\
        \hline

    \end{tabular}%
    }
    \label{tab:2}
\end{table}

\subsubsection{Reconstruction of time series}

Commonly used time series reconstruction methods can be categorized into two main groups. One group involves interpolation and smoothing, which accounts for noise (e.g., clouds, shadows, and snow), including Savitzky-Golay algorithm \citep{Cai18}, Whittaker smoother \citep{EilersPaulHC17}, Fourier transformation algorithm \citep{Khanal20}, and the linear-fit averaging smoother \citep{Khanal20}. Another group stems from resampling, which addresses noise as well as inconsistent observation frequency across pixels due to the overlap along and across orbits. To convert irregular temporal sampling into evenly spaced time intervals, one can aggregate date periods of a certain length \citep{Yang24, YumiaoW23} or perform linear resampling using a certain date interval \citep{Xu20a, Griffiths19, Xuan23}. The hybrid method applies an additional smoothing step to the resampled time series, preserving temporal trends while reducing noise and outliers \citep{Cai18, Feng23, Xuan23, YumiaoW23}. While cloud contamination limits the utility of raw optical imagery for visual interpretation \citep{LillesandThomas15}, some studies argue that noise-resilient architectures, such as Recurrent Neural Network (RNN) and Transformer, can still achieve comparable classification accuracy without preprocessing \citep{MarcRußwurm20, Perich23, TangPengfei24}, thereby supporting the value of the raw image time series.

Selecting temporal windows from the full-season data also impacts the reconstructed time series. A common practice is to select a fixed temporal window aligned with feature importance and typical crop growing season, usually corresponding to the crucial growing periods of dominant crops \citep{Li24, YumiaoW23, JinfanXu21}. Alternatively, the window can be shifted to align with field-specific growing seasons, enabling the dynamic temporal windows \citep{Yang24, NyborgJoachim22, KernerHannahR22}. Empirical comparisons between fixed and dynamic approaches remain limited, and no consensus has been reached on optimal strategies. Nevertheless, no systematic evaluation has been conducted to compare these preprocessing approaches on large-scale crop mapping.

\subsubsection{Feature selection strategies}
Beyond time series-based features, incorporating complementary variables such as empirical indices, statistical features, and multi-source data can enrich the input and improve classification performance. Conventional approaches augmented optical variables with empirical vegetation indices, such as the Normalized Difference Vegetation Index (NDVI; \citet{Xuan23}). Some studies also integrated statistical metrics into the inputs, including temporal mean, variance, and percentiles, to enhance class separability for ML models \citep{JoséMPeñaBarragán11}. Integrating multi-source optical imagery and synthetic aperture radar (SAR) enhances the temporal density and reliability of the time series. Additional optical sources increase the number of usable observations \citep{Li24, YangC21, Griffiths19}, while SAR can penetrate clouds, allowing for consistent Earth observation even under frequent cloud cover \citep{LiuChenfang24, LillesandThomas15}. Some studies also fuse topographic and meteorological variables such as precipitation, temperature, and growing degree days \citep{LihengZhong14, Ravirathinam24}. More recently, self-supervised learning based on DL models has enabled representation extraction from spectral-temporal data \citep{XuYijia24}, followed by site-specific fine-tuning. Although DL can infer high-level features, the interpretability of these features remains a challenge \citep{WenfangZ24}. Most existing work assumes a fixed sample size when selecting features. The combined effects of complementary variable types and varying sample sizes on data richness and large-scale crop classification remain largely underexplored.

\subsection{Pixel-wise crop classification approaches} \label{sec:intro_model}
A variety of ML and DL algorithms have been applied to pixel-wise crop classification and transfer learning. This section reviews key methods that have been applied in time series-based fully supervised learning (Table~\ref{tab:3}) and transfer learning (Table~\ref{tab:4}) for large-scale crop mapping.

\begin{table}[ht]
    \centering
    \caption{Types of pixel-wise crop classification models}
    \renewcommand{\arraystretch}{1.5}
    {\footnotesize
    \begin{tabular}{
        >{\arraybackslash}m{4cm}  >{\arraybackslash}m{6cm} >{\arraybackslash}m{5cm}
    }
        \hline
        \textbf{Model Type} & \textbf{Learning Focus} & \textbf{Example} \\
        \hline
        \multirow{3}{*}{Machine learning models} 
        & \multirow{3}{6.5cm}{Nonlinear projection} 
        & RF \citep{Griffiths19, Xuan23, Li24} \\
        & & SVM \citep{Ndikumana18} \\
        & & KNN \citep{Ndikumana18} \\
        \hline
        \multirow{5}{*}{Recurrent Neural Networks} 
        & \multirow{5}{6.5cm}{Sequential spectral-temporal dependencies} 
        & LSTM \citep{MarcRußwurm20} \\
        & & GRU \citep{Ndikumana18} \\
        & & Bi-LSTM \citep{WangX24} \\
        & & AtBi-GRU \citep{Feng23} \\
        & & AtBi-LSTM \citep{Xu20a} \\
        \hline
        \multirow{2}{*}{Convolutional Neural Networks} 
        & \multirow{2}{6.5cm}{Local spectral-temporal dependencies} 
        & Pixel-level ConvStar \citep{Selea23} \\
        & & Pixel-level Conv1D \citep{WangX24} \\
        \hline
        Transformer & Global and long-term dependencies & Transformer \citep{JinfanXu21} \\
        \hline
    \end{tabular}%
    }
    \label{tab:3}
\end{table}

\subsubsection{Supervised learning models}
Crop mapping models predominantly rely on supervised learning in data-rich regions, where ML models learns by maximizing class separation, and DL algorithms learns by loss propagation to approach the ground truth labels. Traditional ML methods like Random Forest (RF), Support Vector Machines (SVMs), and DL models, such as Multilayer Perceptron (MLP), excel at capturing nonlinear relationships \citep{Xuan23, KokZhiHong21, KussulNataliia15}, but they cannot autonomously extract temporal dependencies from time-series data \citep{Xu20a}. 

In contrast, sequential DL models such as RNNs, Long Short-Term Memory (LSTM), and Gated Recurrent Units (GRU) leverage temporal context by propagating hidden states through time \citep{Selea23, JinfanXu21, ElmanJeffreyL90}. Each node receives information from both the current time step and previous observations, enabling the model to capture spectral-temporal variation patterns of crop growth. LSTM and GRU architectures incorporate gating mechanisms to manage long-term dependencies effectively, proving useful for modeling crop development \citep{MarcRußwurm20, Ndikumana18, Zhong19}. Another DL model architecture is Convolutional Neural Network (CNN), which excels in patch-based classification. Its one-dimensional CNN architecture (1DCNN) also performs well in time series-based pixel-wise crop classification by sliding the kernel along the temporal dimension to capture local spectral-temporal dependencies \citep{WangX24, PhamVuDong24}. In contrast to RNNs and CNNs, Transformers learn global dependencies rather than local or sequential dependencies based on a self-attention mechanism \citep{JinfanXu21}. The Transformer architecture has demonstrated strong performance in patch-based crop mapping tasks \citep{FanLingling24, MarcRußwurm20}. Furthermore, hybrid architectures such as the attention-enhanced bidirectional RNN (AtBi-RNN) integrate the attention mechanism to enhance the discriminative capability of sequential architectures \citep{Feng23, Xu20a}. However, DL models do not universally surpass traditional ML methods. Simpler algorithms, such as RF, are still competitive in smaller or less diverse dataset scenarios \citep{Zhong19, LiZiming24}.

\subsubsection{Transfer learning approaches}

\begin{table}[ht]
    \centering
    \caption{Summary of transfer learning approaches and label requirements in crop mapping}
    {\footnotesize
    \begin{threeparttable}
        \begin{tabular}{m{1.5cm} >{\centering\arraybackslash}m{1cm} >{\centering\arraybackslash}m{1cm} >{\centering\arraybackslash}m{1cm} m{7cm} >{\centering\arraybackslash}m{2cm}}
            \hline
            \textbf{Approach}\tnote{1}  & \textbf{Source label} & \textbf{Pre-trained model} & \textbf{Target label}\tnote{2} & \textbf{Description} & \textbf{Example}\tnote{3} \\
            \hline
           Direct transfer & N & Y & N & Relies on source-site pre-trained model generalizability to perform well on the target domain. & \citep{Xu20a} \\ \\
           MTL & Y & N & Optional & Trains models on related source-site tasks that share feature representation \citep{caruana97}; the trained model's generalizability achieves transfer. & \citep{zhang2021survey}	\\ \\
           Fine-tuning & N & Y & Y & Adapts a pre-trained model using a small number of labeled samples from the target domain. & \citep{ArturN21} \\ \\
           UDA & Y & N & N & Aligns feature distributions between labeled source data and unlabeled target data during training. & \citep{YumiaoW23}\\ \\
           SFUDA & N & Y & N & Adapts a pre-trained model to new target domains using only unlabeled target data; source data is not required at adaptation stage \citep{BoudiafMalik22}. & \citep{SinaMohammadi24}\\ \\
           Few-shot learning & Y & Y & Y & Uses labeled source-site data as the support dataset and a few labeled target-site samples as the query to adapt a pre-trained model, typically without requiring strict task alignment. & \citep{mohammadi2024}\\
            \hline
        \end{tabular}%
    \begin{tablenotes}
        \item[1] Abbreviations: UDA, unsupervised domain adaptation; SFUDA, source-free unsupervised domain adaptation; MTL, multi-task learning; Y: required; N: unneeded.
        \item[2] The availability of pre-trained models and labels indicates their usefulness in transfer or adaptation. Test set labels used solely for evaluation are excluded.
        \item[3] Some examples are patch-based crop mapping.
    \end{tablenotes}
    \end{threeparttable}
    }
    \label{tab:4}
\end{table}

Some DL architectures possess strong direct-transfer capabilities \citep{Xu20a}, and transfer learning techniques can adapt the model to better align the characteristics of the target domain \citep{PanSinnoJialin10}. In particular, transfer learning enables the model to transfer knowledge from data-rich domains to data-poor domains, thereby achieving crop classification in the target domain with minimal (supervised transfer learning) or no labeled samples (unsupervised transfer learning).

The most widely used supervised strategy is fine-tuning, which adapts pre-trained models using a limited number of labeled target samples. This typically involves adjusting frozen layers and tuning parameters, such as learning rates and batch sizes \citep{MilosPandzic24, YangLingbo23, ArturN21, GadirajuKrishnaKarthik20}. Additionally, multi-task learning (MTL), although not specifically intended for transfer learning, trains models on related tasks that share feature representations \citep{caruana97}, thereby enhancing model generalizability and potentially contributing to transferability. For instance, in the temporal transfer scenario, the MTL approach outperforms single-site transfer in large-scale crop mapping for its improved handling of spatial variability \citep{LinZhixian22}. Moreover, the few-shot learning (FSL) method, primarily used in image classification and natural language processing, is also emerging in crop mapping. \citet{mohammadi2024} explored eight FSL methods to map infrequent crops as well as diverse and complex agricultural areas. They suggested pre-training on the label-rich source dataset that is more similar to the target dataset could lead to better results. While another study on classifying different land use types across continents shows that the FSL model underperforms in similar source and target domain transfer settings, but outperforms the direct transfer models and fine-tuning when the target site and source site have a large domain shift \citep{RusswurmMarc20}.

Unsupervised domain adaptation (UDA) is the second most widely used transfer learning approach in environmental remote sensing \citep{MaYuchi24}. By aligning feature distributions between source and target domains, UDA improves model transferability in the absence of labeled target data \citep{GaninYaroslav15}. Frameworks such as Domain-Adversarial Neural Networks (DANN) have shown strong performance across spatial, temporal, and spatiotemporal transfer tasks \citep{YumiaoW23}. CropSTGAN further enhances generalization by generating synthetic target-like features to mitigate domain discrepancies \citep{WangY24}. More recently, source-free UDA methods have emerged, which adapt pre-trained models using only unlabeled target data without requiring access to the source domain \citep{SinaMohammadi24}.

\subsection{Identified research gaps and opportunities} \label{sec:Intro_gaps}

Despite significant advancements in pixel-wise crop type mapping, researchers still face challenges in selecting effective workflows amid the growing diversity of methods. Unlike localized crop classification, large-scale crop mapping requires harmonized preprocessing and model architectures that are consistent, scalable, and adaptable across diverse regions. Comprehensive benchmarking of core components in conventional train-from-scratch workflows remains limited \citep{JoshiAbhasha23}, let alone a systematic evaluation of transfer learning approaches for large-scale crop mapping. 

Recent reviews have outlined datasets and DL applications in crop mapping and yield prediction \citep{JoshiAbhasha23, TeixeiraIgor23}, but they fall short of evaluating DL models through implementation or direct comparison. As such, they do not provide actionable guidance on optimal preprocessing approaches and model design. Among studies with experimental support, \citet{TangPengfei24} and \citet{Feng23} evaluated multiple pixel-wise and patch-based DL models using time series imagery at the region level. These studies, while informative for model selection, do not explore the interaction between different preprocessing strategies and model choices under realistic, variable conditions.
 
Compounding these gaps is the persistent lack of in-situ data across many regions \citep{SafonovaAnastasiia23}. Training accurate models from scratch typically requires large quantities of ground truth samples, which are often unavailable or infeasible to collect in under-surveyed areas. Transfer learning presents a viable alternative, enabling model adaptation across diverse domains. \citet{MaYuchi24} provided a comprehensive overview of transfer learning in environmental remote sensing. They indicate effectiveness depends on the careful selection of source-target domain pairs, high-quality benchmark datasets, and well-matched model architectures \citep{MaYuchi24}. However, implementation-based comparisons and performance evaluations of these techniques were beyond the scope of their study. \citet{ArturN21} conducted experimental studies on transfer learning for crop mapping, focusing solely on configurations within fine-tuning but without exploring the impacts of preprocessing or domain variability. Similarly, \citet{SinaMohammadi24} investigated transfer learning for crop type classification using UDA; however, their study was restricted to UDA configurations and did not compare it with supervised fine-tuning. Collectively, these studies represent significant progress but remain limited in transfer learning-based crop mapping.

To date, no review has systematically compared the foundational components of both conventional and transfer-based crop mapping workflows. This study presents the first comprehensive review of large-scale, pixel-wise crop mapping methods supported by systematic experiments. Building on this foundation, the study aims to:
 
\begin{enumerate}
\itemsep=0pt
\item Identify optimal preprocessing methods and model architectures for large-scale, pixel-wise crop mapping.
\item Evaluate predominant transfer learning approaches for different magnitudes of domain shift.
\item Assess the effects of sample size and variable complementarity on crop mapping performance.
\end{enumerate} 

We achieved these objectives through structured evaluation of time series reconstruction approaches (Section~\ref{sec:analysis_result}), supervised crop classification workflows (Section~\ref{sec:cropclassification}), workflow direct transferability (Section~\ref{sec:directtransferability}), and transfer learning approaches (Section~\ref{sec:improvetransfer}). These objectives and evaluations offer guidelines for creating accurate, scalable, and transferable crop mapping workflows for both data-rich and scarce regions. We used Landsat data for the case studies due to its extensive historical archive and consistent acquisitions, which make it the only source for generating historic crop maps. While Sentinel-2 provides higher temporal frequency and spatial resolution, the techniques discussed in our paper are also applicable to Sentinel-2 data.

\section{Materials and methods}

\subsection{Study area and data}
\begin{figure}
	\centering
		\includegraphics[scale=0.4]{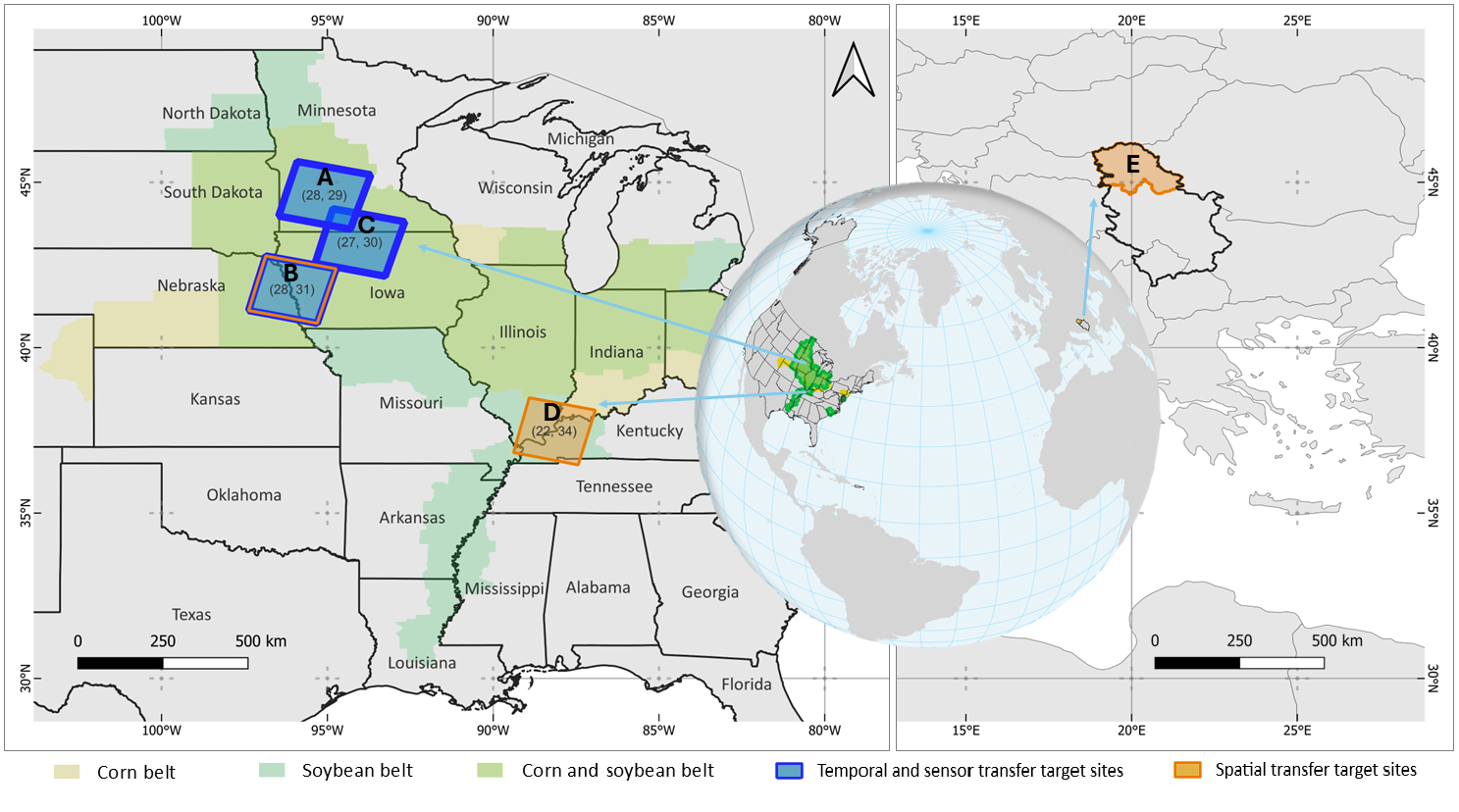}
	\caption{Geographic distribution of five study sites. Sites A, B, and C are located within the U.S. Corn and Soybean Belt and serve as both source and target sites in temporal and sensor transfer scenarios, denoted by blue borders. In the spatial transfer scenario, Site A is designated as the source site, while Sites B, D, and E are the target sites, denoted by orange borders. Site D lies in a transitional zone between the core Corn and Soybean Belt and a predominantly soybean-growing region. Site E is situated in Vojvodina, Serbia, a major agricultural hub in Southeastern Europe.}
	\label{fig:2}
\end{figure}

\begin{figure}
	\centering
		\includegraphics[scale=0.35]{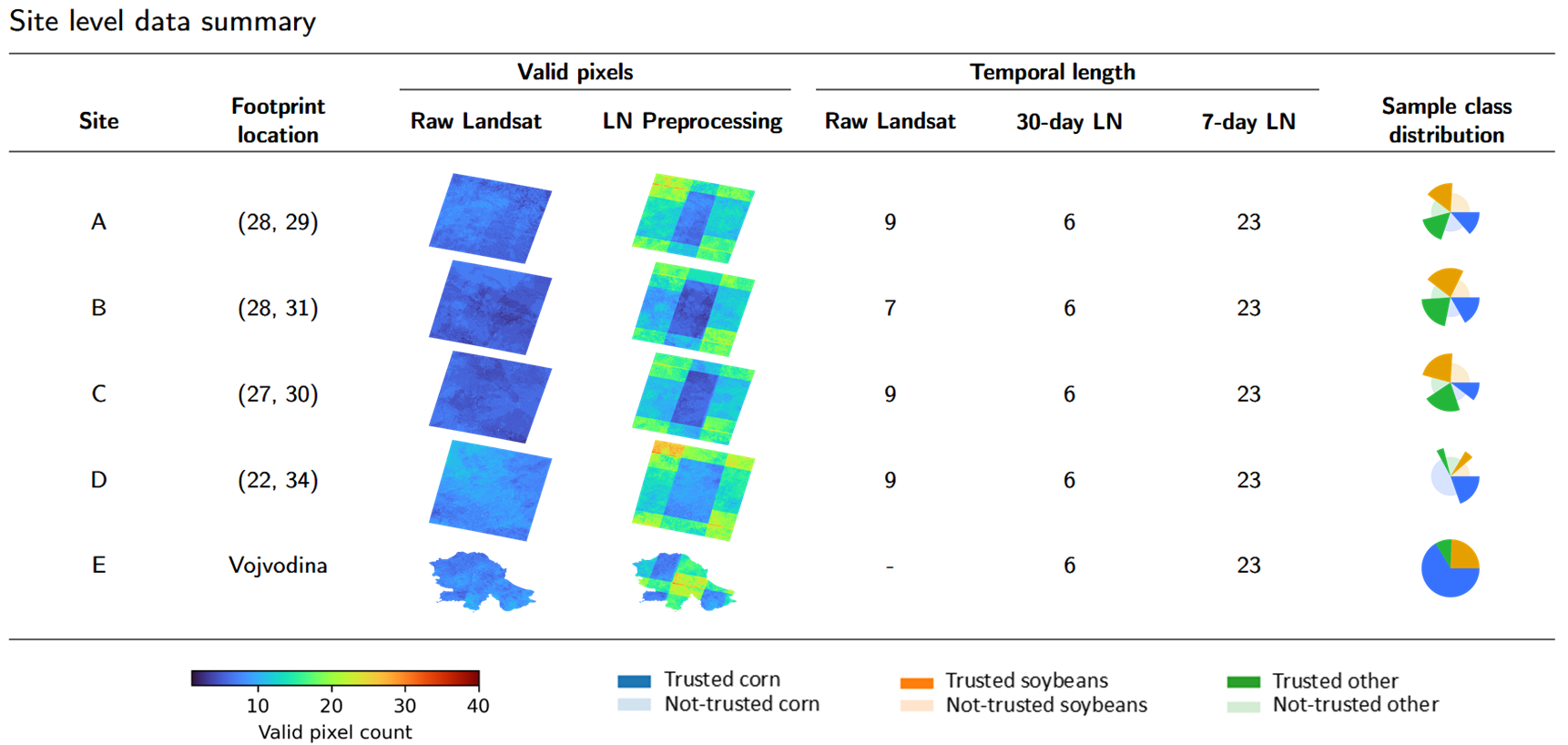}
        \caption{Geographic identifiers, valid pixel distribution, time series length, and crop-specific sample composition for each study site from April 1 to October 1, 2023. The footprint location column lists the geographic identifiers for each site, including Landsat path/row for U.S. regions and the province name for the Serbian site. Valid pixel count refers to the number of pixels with reliable surface reflectance in each observation. Columns 3-4 display the spatial distribution of valid pixels under the raw Landsat and LN-based (linear resampling) preprocessing methods, respectively, with warmer colors indicating a higher number of valid observations. Columns 5-7 report the temporal length resulting from each preprocessing strategy. The final column illustrates the class distribution of reference labels. Bright colors indicate trusted labels used for model training, while pale colors denote non-trusted labels excluded from the experiments.}
	\label{fig:3}
\end{figure}

\subsubsection{Study area}

This study focuses on major agricultural regions in the U.S. Soybean Belt and the Vojvodina region of Serbia (Fig.~\ref{fig:2}). We selected five study sites to represent large-scale, intensively cultivated regions with varied meteorological conditions, field structures, and cropping systems. 

Sites A-D correspond to four Landsat 8 footprints across key corn-soybean producing states in the U.S. Sites A-C, in the northern Corn and Soybean Belt, encompass Minnesota, Iowa, South Dakota, and Nebraska, while Site D spans the intersection of Illinois, Indiana, Kentucky, Tennessee, and Missouri. These regions predominantly fall within a hot-summer humid continental climate (Dfa), with Site D located near the transition to a humid subtropical zone (Cfa; \citet{KottekMarkus06}). Large, contiguous fields are the characteristics of these sites, with average farm sizes in 2022 ranging from 605 ha in South Dakota and 400 ha in Nebraska to 69-307 ha in other states \citep{usdaStateStats}. Most croplands are rainfed, with only 2.3-6\% irrigated, while in Nebraska, over 34\% of agricultural land is irrigated \citep{usdaStateStats}. Corn-soybean rotation is the primary cropping system in Sites A-C and a significant component of agriculture in Site D. Corn is typically planted from April to May and harvested from September to October, while soybean follows a slightly later cycle, with harvest extending through November \citep{nass2010field}.

Site E is located within the fertile Pannonian Plain, a major agricultural hub in Serbia and a key production region in Southeastern Europe. The region experiences a fully humid temperate climate with hot summers (Cfa; \citet{Gordan24, GavrilovMilivoj15}), and annual precipitation between 540-660 mm \citep{TosicIvana13, GavrilovMilivoj15}. In contrast to the other sites, Site E features smaller, more fragmented farms averaging 3.6 ha \citep{novkovic2013} and a diverse cropping system with two distinct growing seasons. Spring crops, including corn, soybeans, sunflowers, and sugar beets, are cultivated from March to November \citep{RadulovicMirjana25}. Winter crops such as wheat, barley, and oilseed rape are sown in autumn and harvested by early summer \citep{usda2019serbia}. The majority of farmland in Vojvodina is rainfed, with only \textasciitilde3\% under irrigation \citep{Radulovic25, RadulovicMirjana23, serbiaWater2022}. 

These five sites, each covering an area of approximately 185 km by 170 km, enable consistent crop classification method comparisons across diverse regions and support spatial transfer experiments, from low-shift domain pairs to cross-continental settings. The geographic identifiers of study sites are presented in Fig.~\ref{fig:3}. We focus on corn and soybean, the two common crops across all sites, while all other land cover types are grouped into the "other" class.

\subsubsection{Satellite data}

Landsat 8 (LC08) Level 2, Collection 2, Tier 1 surface reflectance imagery from 2022 and 2023 served as the primary input for time series reconstruction and classification \citep{usgs_landsat8}. With a 30-meter spatial resolution and a 16-day revisit cycle, Landsat 8 provides high spatial coverage and remains widely used for large-scale crop mapping and long-term agricultural monitoring. Six spectral bands, blue, green, red, near-infrared (NIR), and two shortwave infrared bands (SWIR1 and SWIR2), acquired from Operational Land Imager (OLI), were used to represent core reflectance properties. Cloud and shadow contamination were removed using the built-in quality assessment (QA\_PIXEL) band, thus retaining the valid pixels for each observation. Due to varying cloud cover and orbital overlaps, the number of valid observations varies across regions. Fig.~\ref{fig:3} shows the spatial distribution of valid pixel counts from April to October 2023, with Sites B and C exhibiting higher contamination levels than the other sites.

Additionally, Landsat 7 (LE07) Level 2, Collection 2, Tier 1 from 2022 were used to assess the transferability of trained models in the sensor transfer scenario. Images from Landsat 7 share similar spatial resolution and revisit frequency with Landsat 8 \citep{usgs_landsat7}. But six spectral bands were acquired from the different sensor, the Enhanced Thematic Mapper Plus (ETM+).

To expand beyond the six Landsat optical bands, we also derived vegetation and water indices (VIs) and incorporated SAR data, which were later analyzed in relation to sample size and variable complementarity (see Section~\ref{sec:samplesizevariables}). VIs, including NDVI, Enhanced Vegetation Index (EVI), Green Chlorophyll Vegetation Index (GCVI), Land Surface Water Index (LSWI), Normalized Difference Water Index (NDWI), and Normalized Difference Turbidity Index (NDTI), capture vegetation greenness, canopy structure, surface moisture, and brightness characteristics (see Section~\ref{sec:vis}), while SAR time series contribute complementary information related to surface structure and moisture conditions. Sentinel-1 (S1) Ground Range Detected (GRD) radar products were collected during the same period as the corresponding spectral data \citep{esa_sentinel1}. The S1 GRD data have a spatial resolution of 10 meters and a revisit cycle of six days. Images acquired in Interferometric Wide Swath (IW) mode were processed in GEE using denoising and smoothing techniques. The polarization backscatter coefficients, VV and VH, and the ratio (VH/VV) were used as SAR variables \citep{SchlundMichael20}.

Since classification experiments were based on individual sample coordinates, all pixel values from processed optical and radar images were extracted at each sample location and organized into multivariate time series.

\subsubsection{Trusted labels}

Reliable crop labels are crucial for supervised classification and evaluation of transfer learning. For U.S. sites (A-D), trusted labels were derived from the USDA, National Agricultural Statistics Service (NASS) Cropland Data Layer (CDL), which provides annual 30-meter crop type classifications \citep{BoryanClaire11}. Trusted pixels contribute to enhancing label reliability compared to the original CDL (Appendix~\ref{sec:trustedvsCDL}). We implemented multi-year filtering to identify the trusted pixels for corn and soybean, following \citet{RahmanMdShahinoor19}. A corresponding trusted ratio was then developed to downsample the "other" class and ensure representative sampling.

Specifically, one million random points were generated per study site, and CDL label sequences were extracted for eight years, defined as the study year and the seven preceding years (e.g., 2016-2023 for a 2023 study year). Corn and soybean pixels for the study year were retained if their CDL crop labels remained consistent throughout this eight-year sequence. Consistency was based on either uninterrupted single-crop sequences (e.g., continuous corn or soybean) or alternating crop patterns such as "X-S-X-S-X-S-X-S" or "X-C-X-C-X-C-X-C", where C represents corn, S represents soybean, and X represents types different from either S or C. 

The "other" class includes non-target crops and non-crops. We applied the designed trusted ratio \( p_{\text{trusted}} \) (Eq.~\eqref{eq:trusted_ratio}), defined as the number of trusted corn and soybean pixels relative to their original counts in the CDL data, to downsample the "other" pixels randomly. This approach ensured that the final corn, soybean, and "other" trusted labels maintained the relative distribution of sample classes observed in the original one million CDL samples. For the year 2023, the resulting trusted ratios were 0.45 (Site A), 0.59 (B), 0.53 (C), and 0.30 (D), yielding approximately 440,000, 590,000, 530,000, and 290,000 trusted labels, respectively. Sample distribution in Fig.~\ref{fig:3} indicates that sites A and B exhibited balanced class distributions, Site C showed stable corn-soybean rotation, and Site D exhibited higher crop diversity.

\begin{equation}
n_{\text{trusted}}^{\text{other}} = p_{\text{trusted}} \cdot n_{\text{CDL}}^{\text{other}}, \quad
p_{\text{trusted}} = \frac{n_{\text{trusted}}^{\text{corn}} + n_{\text{trusted}}^{\text{soy}}}{n_{\text{CDL}}^{\text{corn}} + n_{\text{CDL}}^{\text{soy}}} 
\label{eq:trusted_ratio}
\end{equation}

\noindent where 
\( p_{\text{trusted}} \): trusted ratio. 
\( n_{\text{trusted}}^{\text{corn}} \), \( n_{\text{trusted}}^{\text{soy}} \), \( n_{\text{trusted}}^{\text{other}} \): number of trusted pixels in the corn, soybean, and "other" classes, respectively. 
\( n_{\text{CDL}}^{\text{corn}} \), \( n_{\text{CDL}}^{\text{soy}} \), \( n_{\text{CDL}}^{\text{other}} \): number of original CDL-labeled pixels in the corn, soybean, and "other" classes, respectively.

At Site E, trusted labels were obtained from BioSense Institute's high-quality ground truth dataset \citep{MilosPandzic24, ZivaljevicBranislav24}, which comprises 13,000 ground samples from 2,000 field-validated parcels collected during the 2023 growing season. The original labels were reclassified as corn, soybeans, and others to keep consistency with the U.S. sites.

\subsection{Experiment design}

\begin{figure}
	\centering
		\includegraphics[scale=0.5]{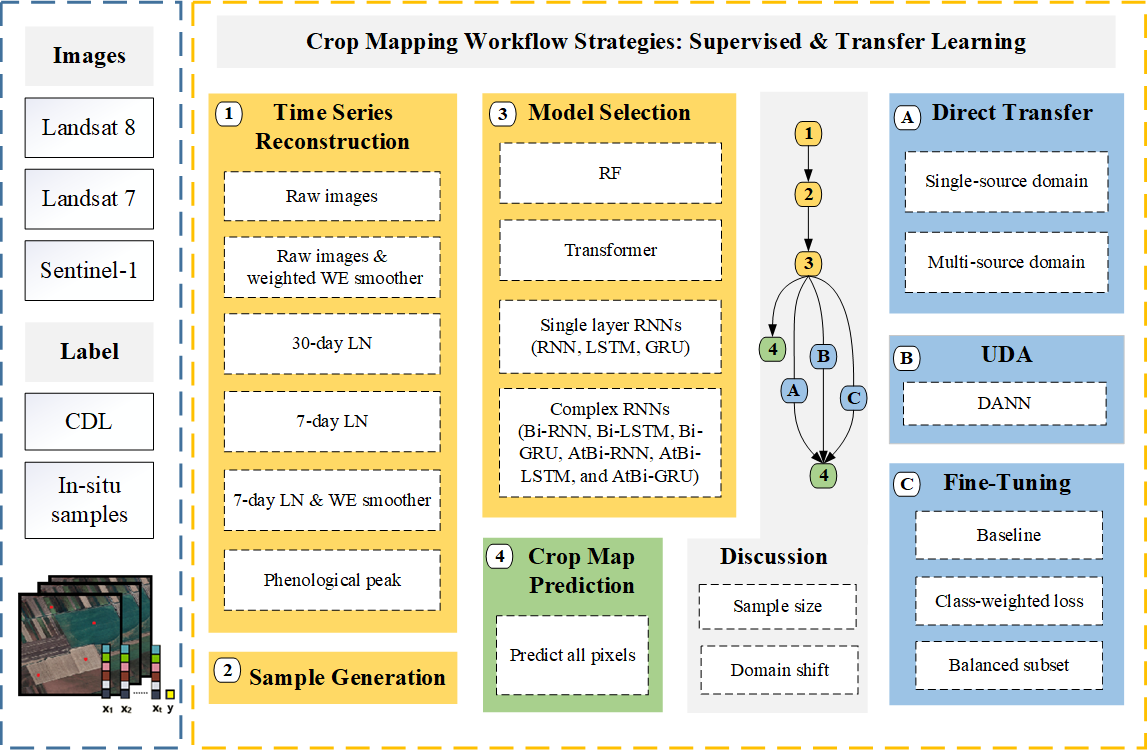}
    \caption{Experimental design overview for evaluating pixel-wise crop mapping workflows. Crop maps were produced either via fully supervised learning or transfer learning. The supervised workflow comprises sequential modules: (1) time series reconstruction, (2) sample generation, (3) model selection, and (4) crop map prediction. Within each module, multiple widely used approaches were evaluated to identify optimal practices. Additionally, the joint impacts of sample size and complementary variables were assessed to guide the selection between fully supervised and transferable approaches. Transferable workflows included (A) direct model transfer, (B) unsupervised domain adaptation (UDA; specifically, DANN), and (C) supervised fine-tuning. Primary experiments utilized Landsat 8 data, while Landsat 7 and Sentinel-1 imagery supported assessments of transferability and data fusion. Crop labels (corn, soybean, and "other") originated from CDL trusted pixels in the U.S. and field surveys in Serbia. Abbreviations: CDL, Cropland Data Layer; WE, Whittaker-Eilers; LN, linear resampling; UDA, unsupervised domain adaptation; DANN, Domain-Adversarial Neural Network.}
	\label{fig:5}
\end{figure}

This study systematically evaluated pixel-wise crop mapping workflows through a series of modular, comparative experiments. For fully supervised crop mapping, the workflow consists of (1) satellite image time series reconstruction, (2) trusted sample generation, (3) model selection, and (4) crop map prediction (Fig.~\ref{fig:5}). Additionally, we discussed the combined impact of sample size and complementary variables on crop classification. Transferable workflows included (A) direct model transfer, (B) UDA (specifically, DANN), and (C) supervised fine-tuning. Table~\ref{tab:5} summarizes labeled sample sizes, variable types, and data partitions used across all experiments, organized by source and target sites applications. 

First, six satellite image-based preprocessing methods were compared, including raw imagery \citep{Li24}, weighted Whittaker-Eilers (WE) smoothing \citep{Whittaker92}, 7-day and 30-day linear resampling \citep{Xu20a}, 7-day linear resampling with WE smoothing \citep{Xuan23}, and 7-day linear resampling centered on the phenological peak. 

Eleven models were evaluated, spanning classical ML, RNNs, and attention-based architectures. The classical baseline was RF \citep{Breiman01}. Sequential models encompassed vanilla RNN \citep{ElmanJeffreyL90}, LSTM \citep{Hochreiter97}, GRU \citep{ChoKyunghyun14}, and their bidirectional variants (Bi-RNN, Bi-LSTM, Bi-GRU; \citet{SchusterMike97}). The Transformer was selected for its attention-driven architecture \citep{JinfanXu21, Vaswani17}, which is effective in capturing long-term spectral-temporal patterns. We also examined attention-enhanced RNNs, including Attention-Bi-RNN (AtBi-RNN), Attention-Bi-LSTM (AtBi-LSTM), and Attention-Bi-GRU (AtBi-GRU), based on the mechanism proposed by \citet{BahdanauDzmitry14}.

Based on the optimal supervised crop classification workflow, sample size and variable combination experiments were conducted. Specifically, four variable types--spectral bands alone, combined with VIs, SAR, or both--were assessed across six sample size levels. 

Three predominant transfer learning techniques consist of model direct transfer, which was evaluated across spatial (different geographic regions), temporal (different years), and sensor (Landsat 8 to Landsat 7) domains, DANN \citep{GaninYaroslav15}, and fine-tuning, the latter two assessed under varying spatial shift conditions.

All experiments were conducted using Python 3.9.18 with PyTorch 1.11.0, running on an Ubuntu 20.04 system with two NVIDIA A100-PCIE-40GB GPUs. Through these well-structured experiments, the study provides comprehensive insights into optimal practices for pixel-level crop mapping, emphasizing scalability, accuracy, and transferability.

\begin{table}[ht]
    \centering
    \caption{Labeled sample sizes and data splits used in experiments}
    {\footnotesize
    \begin{threeparttable}
    \begin{tabular}{p{3.5cm} >{\centering\arraybackslash}p{4cm} >{\centering\arraybackslash}p{0.5cm} >{\centering\arraybackslash}p{0.5cm} >{\centering\arraybackslash}p{0.5cm} >{\centering\arraybackslash}p{4.5cm}}
        \hline
        & \textbf{Labeled Data} & \textbf{Train} & \textbf{Val} & \textbf{Test}\tnote{1} & \textbf{Variables} \\
        \textbf{Source Site Experiments} & \textbf{Source Site} &&&& \\
        \hline
        Best preprocessing methods & \texttt{400,000} & \texttt{80\%} & \texttt{10\%} & \texttt{10\%} & \texttt{6 spectral bands} \\
        Best model architectures & \texttt{400,000} & \texttt{80\%} & \texttt{10\%} & \texttt{10\%} & \texttt{6 spectral bands} \\
        \multirow{2}{=}{Sample size \& variables} 
            & \texttt{400,000, 100,000, 50,000, 10,000, 1,000, 500} 
            & \multirow{2}{=}{\centering\texttt{80\%}} & \multirow{2}{=}{\centering\texttt{10\%}} & \multirow{2}{=}{\centering\texttt{10\%}} 
            & \texttt{6 spectral bands, combined with 6 VIs, 3 SAR, or both} \\
        \\
        \textbf{Target Site Experiments}& \textbf{Target Site} &&&& \\
        \hline
        Direct transfer & \texttt{4,000} & \texttt{X} & \texttt{X} & \texttt{4,000} & \texttt{6 spectral bands} \\
        Fine-tuning & \texttt{8,000} & \texttt{3,000} & \texttt{1,000} & \texttt{4,000} & \texttt{6 spectral bands} \\
        UDA\tnote{2} & \texttt{4,000} & \texttt{X} & \texttt{X} & \texttt{4,000} & \texttt{6 spectral bands} \\
        \hline
    \end{tabular}
    
    \begin{tablenotes}
        \item[1] The labeled test samples were used solely for evaluation.
        \item[2] DANN used a single train-test split. All other experiments used 10-fold cross-validation.
    \end{tablenotes}
    \end{threeparttable}
    }
    \label{tab:5}
\end{table}

\subsection{Satellite image preprocessing methods} \label{sec:preprocessing}

\begin{figure}
	\centering
		\includegraphics[scale=0.6]{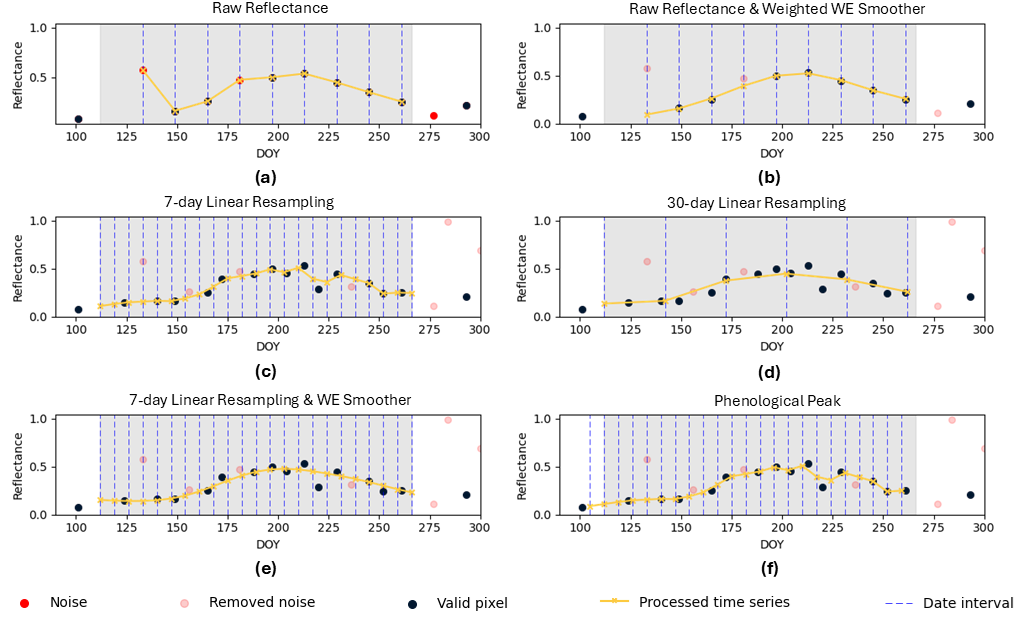}
	\caption{Six image-wide preprocessing methods are exemplified using NIR reflectance from a trusted sample at Site A. The grey background represents the target crop growing season defined in 2023. Observation (dots) selection depends on whether reconstruction relies on native path and row composites (raw methods; (a) and (b)) or all available observations (linear resampling methods; (c)-(f)). The QA\_PIXEL band was used to identify and remove noise (red dots; e.g., clouds, shadows, snow). Yellow curves depict the processed NIR time series, with the x-axis showing day of year (DOY) and the y-axis representing reflectance. (a) Raw reflectance values from the growing season. (b) Weighted WE-smoothed raw data. (c) 7-day linear resampling. (d) 30-day linear resampling. (e) WE-smoothed 7-day linear resampling. (f) 7-day linear resampling focused on phenological peak periods.}
	\label{fig:6}
\end{figure}

Taking the raw imagery as the baseline, we applied five widely used methods to reconstruct the time series and ensure temporal regularity while minimizing noise. These included raw native path and row composite-based (raw-based) methods, as well as all available pixel-based (linear resampling-based) approaches. Fig.~\ref{fig:3} summarizes site-specific valid pixel distribution (before preprocessing) and image-wide uniform temporal length of processed time series. Fig.~\ref{fig:6} illustrates the reconstruction principles using the near-infrared band of a trusted sample from Site A. Fig.~\ref{fig:9} shows reconstructed time series curves for all samples from Site A. Satellite image preprocessing, including denoising and time series reconstruction, is performed through \href{https://code.earthengine.google.com/d5deb3dd74d784d4804249d649699bda}{GEE}.

\subsubsection{Whittaker-Eilers smoother}

The WE smoother promotes temporal smoothness by minimizing differences between consecutive time steps while preserving data fidelity to the observed time series structure \citep{Whittaker92}. This is achieved by optimizing a regularized least-squares function \citep{Xuan23}. We extend the original WE smoother to handle multi-band Landsat 8 imagery and its irregular observation dates by incorporating a time interval-aware weighting matrix, which adjusts weights based on the length of date intervals between observations. This enhancement enables reliable interpolation across gaps caused by cloud cover and inconsistent image acquisition (e.g., Fig.~\ref{fig:6}b illustrates that the Landsat 8 Level-2 product missed an entire scene between DOY 101 and 133 in 2023, despite the satellite's nominal 16-day revisit cycle.) Fig.~\ref{fig:6}b and~\ref{fig:6}e illustrate the difference between weighted and evenly spaced WE smoothing. Sentinel-1 polarization bands were also denoised and smoothed using the weighted WE approach, producing 35 time steps per band corresponding to the original Sentinel-1 acquisition dates in 2023 (Fig.~\ref{fig:9}g).

\subsubsection{Linear resampling}

Linear resampling generates temporally consistent time series by first defining a regular sequence of resampled dates, and then linearly interpolating band values between the nearest valid observations to estimate reflectance at each resample date. Unlike the WE smoother, which reconstructs time series based on valid pixels from individual Landsat acquisition footprints, the linear resampling method utilizes all valid pixels from overlapping revisits across adjacent paths and rows. This approach enhances temporal continuity and captures finer phenological variations. During the target crop growing season (April 21 to September 22; \citet{nass2010field}), two resampling intervals were applied: 7 days, yielding 23 time steps  (Fig.~\ref{fig:6}c), and 30 days, resulting in six time steps  (Fig.~\ref{fig:6}d). Based on the 7-day resampled time series, we implemented a hybrid method combining 7-day linear resampling with additional WE smoothing to suppress outliers and better capture temporal trends  (Fig.~\ref{fig:6}e).

\subsubsection{Phenological peak} \label{sec:phenological_peak}

The phenological peak 7-day linear resampling method defines dynamic temporal windows at the pixel level by analyzing the NDVI time series trajectory. Building on the approach of \citet{KernerHannahR22}, which extracted three discrete phenological markers (green-up, peak greenness, and senescence), our approach upgrades by generating a continuous 23-step time series centered on the NDVI peak, indicating a growing season around the crop greenness summit. Specifically, for each pixel, the phenological peak method identified the peak NDVI within a predefined window (April 21 to September 22) and extracted the surrounding time series by including half the steps before and after the peak (Fig.~\ref{fig:6}f). The resulting five-month, 23-step time series adapts to regional phenological variation and captures spectral variability during the specific growing season at each pixel.

\subsubsection{Evaluating preprocessing methods via Dynamic Time Warping}

Dynamic Time Warping (DTW) is a non-linear sequence alignment algorithm widely used to measure similarity between two temporal sequences \citep{PetitjeanFranois12, CsillikOvidiu19}. In this study, DTW accommodates temporal shifts and unequal sequence lengths, allowing fair comparison of time series that represent the same underlying crop growth dynamics but differ in temporal resolution due to preprocessing methods. Given that the primary goal of preprocessing is to enhance the separability between the classes, particularly in distinguishing between target crops, the DTW analysis focused on two dominant crop types.

The effectiveness of each preprocessing method was quantified by counting the number of spectral bands in which inter-class separability exceeded intra-class separability. Specifically, for each preprocessed time series dataset, 100 corn and 100 soybean samples were randomly selected. For these selected samples, inter-class DTW distances (between corn and soybean samples) and intra-class distances (within corn or within soybean) were calculated for each band. A band was counted as separable if its inter-class distance exceeded both intra-class distances. The total number of such bands served as a summary measure of target crop spectral-temporal separability under each preprocessing method.

\subsection{Pixel-wise classification models} \label{sec:model}

\begin{figure}
	\centering
		\includegraphics[scale=0.3]{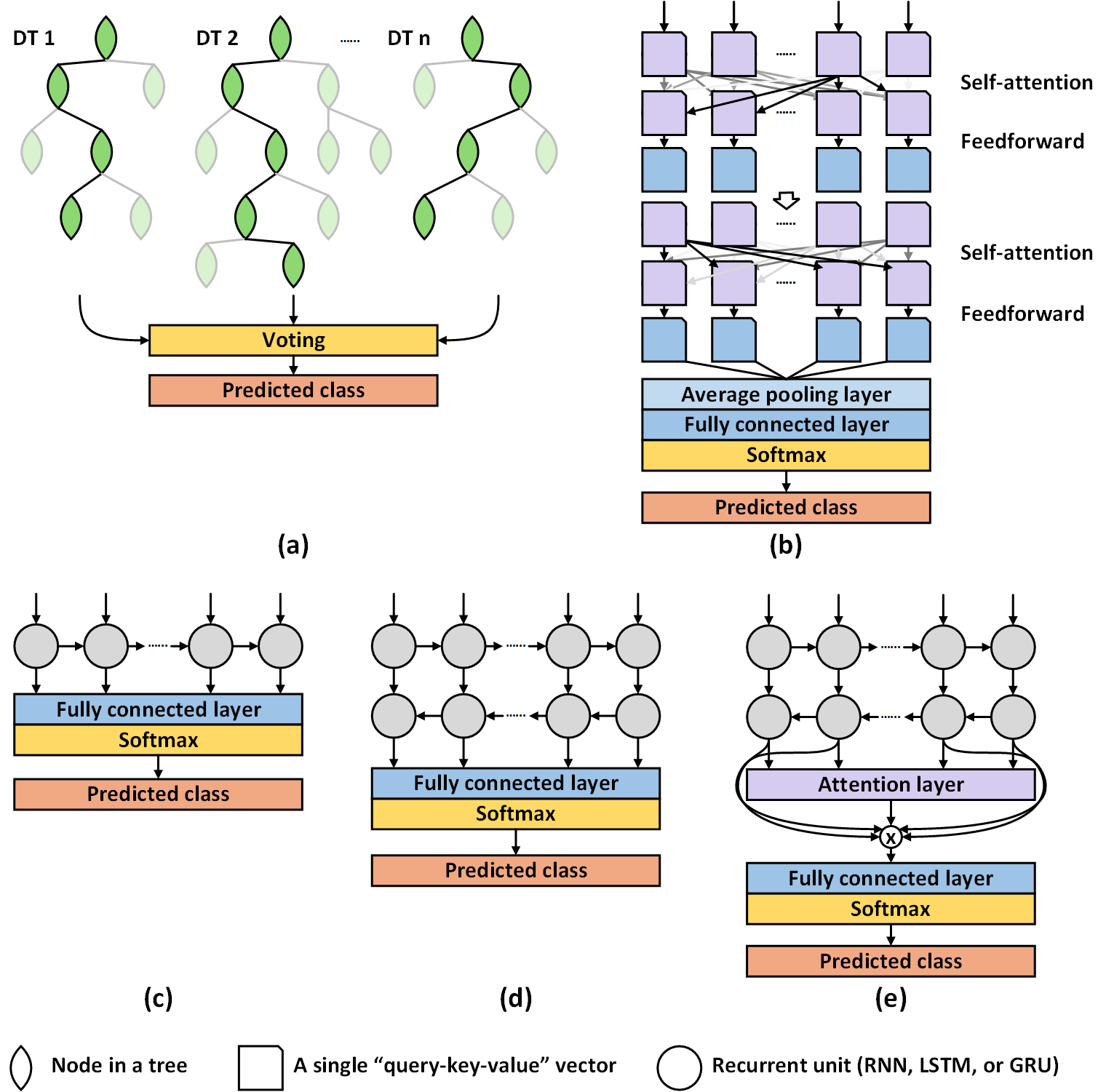}
	\caption{Architecture illustration of pixel-based models. (a) Random Forest. (b) Transformer. (c) Single-layer RNN architecture (RNN, LSTM, and GRU). (d) Bidirectional RNN architecture (Bi-RNN, Bi-LSTM, and Bi-GRU). (e) Bidirectional RNN with an attention layer architecture (AtBi-RNN, AtBi-LSTM, and AtBi-GRU). Abbreviation: DT, decision tree model.}
	\label{fig:7}
\end{figure}

We evaluated eleven pixel-wise classification models based on five architectures, including RF, single-layer RNN, bidirectional RNN, attention-augmented RNN, and Transformer (Fig.~\ref{fig:7}). These models were selected to represent varying architectures for temporal feature learning and a range of architectural complexities. Table~\ref{tab:6} summarizes the hyperparameter settings and training configurations used for the ML and DL models, adapted from prior crop classification studies.

\subsubsection{Random Forest}

The RF algorithm is an ensemble learning method that builds multiple decision trees (DT) using a technique called bootstrap aggregation, or bagging (Fig.~\ref{fig:7}a; \citet{Breiman01}). Each tree is trained on a randomly sampled subset of the training data and input variables, which introduces diversity into the model. During training, each tree splits data at its nodes to maximize class separation. At inference time, predictions from all trees are combined through majority voting to produce the final classification. Despite reduced interpretability due to its ensemble nature, RF is robust to overfitting, computationally efficient, and has demonstrated strong performance across a wide range of classification tasks \citep{Zhong19, LiZiming24}.

\subsubsection{Deep learning model}

DL models designed for sequential data, such as RNNs and Transformers, learn temporal dependencies through backpropagation and iterative loss optimization. RNNs remain the most widely adopted architectures for modeling temporal data in crop mapping and yield prediction \citep{JoshiAbhasha23}, while Transformer-based models are increasingly recognized for their potential to handle complex temporal dependencies, particularly in multimodal data fusion and urban land use classification \citep{LiZiming24}.

RNNs process sequential inputs by updating hidden states at each time step, allowing information from previous observations to influence current predictions (Fig.~\ref{fig:7}c-e). However, the vanilla RNNs' ability to model long-range dependencies is limited by the vanishing gradient problem, which led to the development of more robust variants such as LSTMs and GRUs. LSTMs address the vanishing gradient problem through memory cells and gated operations (input, output, and forget gates; \citet{Hochreiter97}), while GRUs use a simplified structure with reset and update gates to achieve comparable performance with fewer parameters \citep{ChoKyunghyun14, Chung14}. To enhance the RNN's ability to emphasize important temporal features, the hybrid architectures integrated multi-head attention layers into bidirectional RNNs (AtBi-RNNs). These layers assign dynamic weights to different time steps, enabling the model to better capture key crop growth signals \citep{Xu20a, BahdanauDzmitry14}. The attention-weighted features are then passed through a fully connected layer, followed by a softmax function to finally predict crop types.

Transformer enables efficient modeling of long-range temporal dependencies via attention-driven architecture (Fig.~\ref{fig:7}b). In the pixel-wise implementation of the Transformer model, the spectral values of each time step are first embedded and projected into query, key, and value vectors. Self-attention module extracts crucial temporal relationships across the whole sequence of vectors and assigns weights (Fig.~\ref{fig:7}b; \citet{Vaswani17}). The resulting attention-weighted features proceed through a position-wise feedforward network, which further refines the representation at each time step \citep{JinfanXu21}. The Transformer mitigates vanishing gradient issues and supports parallel training, making it particularly well-suited for modeling high-dimensional time series from millions of pixels and complex classification tasks.

\begin{table}[ht]
    \centering
    \caption{Hyperparameters and training parameters for crop mapping models}
    {\footnotesize
    \begin{tabular}{ p{4cm} p{11cm} }
        \hline
        \textbf{Model} & \textbf{Hyperparameters}  \\
        \hline
        RF \citep{Li24} & \texttt{random\_state=100}, \texttt{n\_estimators=500}, \texttt{max\_features=8}  \\
        Single-layer RNNs & \texttt{hidden\_size=256}, \texttt{num\_layers=1}, \texttt{bidirectional}=False  \\
        Bidirectional RNNs  & \texttt{hidden\_size}=256, \texttt{num\_layers}=2, \texttt{bidirectional}=True, \texttt{dropout}=0.2 \\
        Transformer \citep{JinfanXu21} & \texttt{d\_model} = 512, \texttt{nhead} = 8, \texttt{dim\_feedforward} = 256, \texttt{dropout} = 0.2, \texttt{num\_layers} = 2  \\
        Bidirectional RNNs with attention mechanism \citep{Xu20a}  & \texttt{hidden\_size}=256, \texttt{num\_layers}=2, \texttt{bidirectional}=True, \texttt{dropout}=0.2 \\

        \hline
        \textbf{Training Workflow} & \textbf{Training Parameters}   \\
        \hline
        RF & \texttt{n\_jobs}=-1 \\
        DL models & Using the cross-entropy loss function and the Adam optimizer with a learning rate scheduler. \texttt{lr}=0.0005, \texttt{batch\_size}=4,096 \\
        Fine-tuning & Using the cross-entropy loss function and the Adam optimizer with a learning rate scheduler. \texttt{lr}=0.00001, \texttt{batch\_size}=32  \\
        \hline
    \end{tabular}%
    }
    \label{tab:6}
\end{table}

\subsection{Transfer learning technique}

Transfer learning techniques adapt DL models to new target domains, aiming to mitigate model performance degradation with minimal or no target-site labeled data. This study focuses on two predominant transfer learning techniques: fine-tuning and UDA. Fine-tuning adapts a pre-trained model using a small set of target-domain labeled samples. This supervised adaptation is typically performed with a reduced learning rate and fewer training epochs to balance effective adaptation with the risk of overfitting. While certain research suggests freezing the shallow layers to preserve generalizable features \citep{MilosPandzic24}, complete fine-tuning, which involves updating all parameters, is found to be more efficient for transfers across areas with similar land cover characteristics \citep{ArturN21}. Therefore, full model fine-tuning was applied in all experiments. To mitigate class imbalance at specific target sites, additional configurations were incorporated, including undersampling and weighted loss.

\begin{figure}
	\centering
		\includegraphics[scale=0.5]{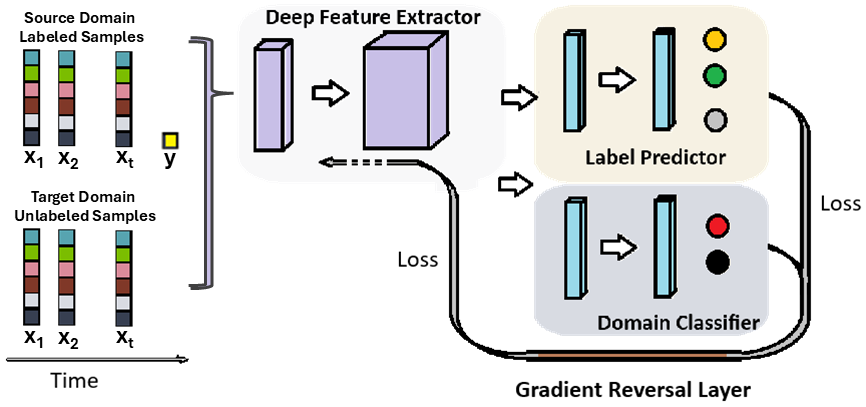}
    \caption{Overview of the DANN framework used for unsupervised domain adaptation. The model processes time series from both source and target domains using a shared deep feature extractor. The resulting feature representations are passed to two branches: a label predictor trained on source labels and a domain classifier trained to distinguish between source and target samples. A gradient reversal layer connects the feature extractor to the domain classifier, inverting its loss to promote the learning of domain-invariant yet class-discriminative features. Cuboids represent deep feature vectors; colored circles illustrate class predictions (yellow for corn, green for soybean, gray for "other") and domain discrimination (red for invariant features, black for domain-variant features).}
	\label{fig:8}
\end{figure}

UDA addresses transfer scenarios where target domain labels are unavailable. We implemented the DANN framework \citep{Ganin16, JunguangJiang22}, which adapts the model to the target domain by learning domain-invariant features \citep{YumiaoW23}. As shown in Fig.~\ref{fig:8}, the input consists of both source-domain labeled data and target-domain unlabeled data. A shared deep feature extractor, typically a DL architecture excluding the final classification layer, processed the inputs to deep features. The resulting feature representations are then forwarded to two branches: a label predictor trained using source domain labels and a domain classifier trained to distinguish between source and target features. A gradient reversal layer connects the feature extractor to the domain classifier, inverting the domain classification loss during backpropagation and thereby guiding the feature extractor to produce domain-invariant features. This setup encourages the model to learn features that are discriminative for crop classification while being robust to domain shifts, such as variations in phenological timing. Training is performed via stochastic gradient descent, jointly optimizing both the classification and domain adversarial losses.

\subsection{Vegetation and water indices} \label{sec:vis}

Six VIs were derived from the surface reflectance (\( \rho\)) of Landsat 8 optical bands (Eq.~\eqref{eq:ndvi}-Eq.~\eqref{eq:ndti}). NDVI, EVI, and GCVI are widely used to estimate chlorophyll content and overall vegetation vigor \citep{ComptonJTucker79, LiuHuiQing95, HueteAR97, AnatolyAGitelson03}. LSWI and NDWI are sensitive to surface and canopy water content, soil moisture, and open water bodies \citep{KChandrasekar10, gao1996ndwi}. In general, higher NDVI, EVI, and GCVI values indicate healthier and denser vegetation, while elevated LSWI and NDWI values reflect increased surface moisture or water presence. NDTI complements these indices by improving the separation of turbid water bodies from vegetated or bare soil areas \citep{lacauxJP07}.

\begin{equation}
    \text{NDVI} = \frac{\rho_{\text{NIR}} - \rho_{\text{RED}}}{\rho_{\text{NIR}} + \rho_{\text{RED}}}
    \label{eq:ndvi}
\end{equation}

\begin{equation}
    \text{EVI} = 2.5 \times \frac{\rho_{\text{NIR}} - \rho_{\text{RED}}}{\rho_{\text{NIR}} + 6\rho_{\text{RED}} - 7.5\rho_{\text{BLUE}} + 1}
    \label{eq:evi}
\end{equation}

\begin{equation}
    \text{GCVI} = \left(\frac{\rho_{\text{NIR}}}{\rho_{\text{GREEN}}}\right) - 1
    \label{eq:gcvi}
\end{equation}

\begin{equation}
    \text{LSWI} = \frac{\rho_{\text{NIR}} - \rho_{\text{SWIR1}}}{\rho_{\text{NIR}} + \rho_{\text{SWIR1}}}
    \label{eq:lswi}
\end{equation}

\begin{equation}
    \text{NDWI} = \frac{\rho_{\text{RED}} - \rho_{\text{SWIR1}}}{\rho_{\text{RED}} + \rho_{\text{SWIR1}}}
    \label{eq:ndwi}
\end{equation}

\begin{equation}
    \text{NDTI} = \frac{\rho_{\text{RED}} - \rho_{\text{GREEN}}}{\rho_{\text{RED}} + \rho_{\text{GREEN}}}
    \label{eq:ndti}
\end{equation}

\section{Results}

\subsection{Analysis of time series reconstruction approaches} \label{sec:analysis_result}

\begin{table}[ht]
\centering
\caption{Number of bands where the DTW distance between two target crops exceeds both intra-class DTW distances, calculated from randomly selected 100 corn and 100 soybean samples.}
    {\footnotesize
    \begin{threeparttable}
        \begin{tabular}{p{8cm}>{\centering\arraybackslash}p{6cm}}
            \toprule
            \textbf{Method} & \textbf{Inter-class>Intra-class Band Count\tnote{1}}\\
            \hline
            Raw reflectance&2\\
            Raw reflectance with weighted WE smoothing &5\\
            7-day linear resampling&5\\
            30-day linear resampling&5\\
            7-day linear resampling with weighted WE smoothing &5\\
            Phenological peak&6\\
            \hline
        \end{tabular}%
    \begin{tablenotes}
        \item[1] The detailed DTW values for the six optical band time series used in this analysis are provided in Appendix~\ref{sec:dtwappendix}.
    \end{tablenotes}
    \end{threeparttable}
    }
\label{tab:7}
\end{table}

\begin{figure}
	\centering
		\includegraphics[scale=0.5]{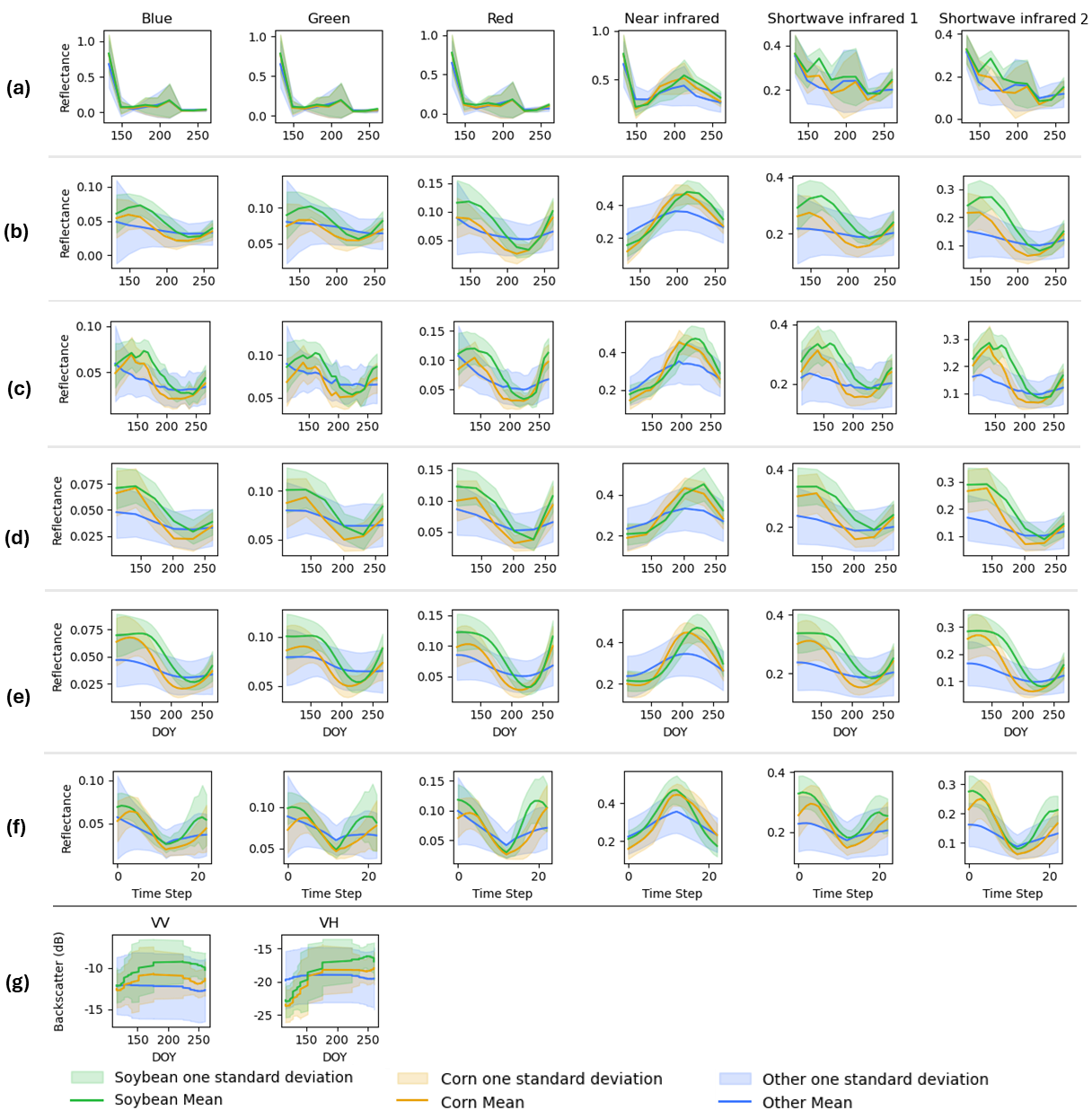}
	\caption{Processed time series of six Landsat 8 optical bands and two Sentinel-1 polarization bands for Site A in 2023. The x-axis shows the time in the format of either DOY or time steps, and the y-axis represents the reflectance value. (a) Raw reflectance values from the growing season. (b) and (g) Weighted WE smoother applied to raw observations. (c) 7-day linear resampling. (d) 30-day linear resampling. (e) WE-smoothed 7-day linear resampling. (f) 7-day linear resampling focused on phenological peak periods.}
	\label{fig:9}
\end{figure}

We evaluated six time-series reconstruction methods by two complementary assessments. First, we performed a visual comparison by plotting the mean and standard deviation curves of six-band time series for all samples across three classes (corn, soybean, and "other") in Site A (Fig.~\ref{fig:9}). Second, we assessed crop separability using DTW-based analysis on randomly selected 100 corn and 100 soybean samples from Site A in 2023. Table~\ref{tab:7} reports the number of bands in which inter-class DTW distances exceeded both intra-class distances, serving as a proxy for spectral-temporal separability of corn and soybeans.

Raw growing-season band values preserved the full spectral details but showed substantial noise, evidenced by extreme reflectance values and high time series variability (Fig.~\ref{fig:9}a). Raw sequences also exhibited poor separability of corn and soybean, as evidenced by only two bands with inter-class DTW values larger than intra-class DTW values (Table~\ref{tab:7}). On the contrary, all preprocessing techniques improved separability to varying extents, as the number of bands that could better separate the target crops increased to five or six. Fig.~\ref{fig:9}b shows the weighted WE smoothing method builds on the raw approach by applying a smoothing algorithm to reconstruct band values at time steps flagged as noise and removed. The weighted WE smoothing effectively reduced noise, stabilized trends, and enhanced the clarity of spectral-temporal patterns.

Unlike raw image-based methods that rely on native path and row composites, linear resampling methods effectively address the limitations of sparse or irregular observations by leveraging all valid acquisitions, including those from overlapping adjacent paths and rows within the study area (Fig.~\ref{fig:9}c and d). The 7-day linear resampling preserves finer spectral details and introduces subtle within-season variability, likely reflecting phenological differences across the large study area. In contrast, the 30-day linear resampling compresses spectral dynamics into six coarse monthly intervals, potentially oversmoothing critical crop growth patterns. Applying WE smoothing to the 7-day resampled data (Fig.~\ref{fig:9}e) reduced residual noise and within-season variability, yielding more visually consistent and homogeneous spectral-temporal patterns across the study area.

The phenological peak method, which extracts 23 time steps centered on each pixel's NDVI maximum, preserved ample within-season variability and improved spectral intra-class homogeneity across Site A (Fig.~\ref{fig:9}f) compared to fixed-window methods (Fig.~\ref{fig:9}c). This is also implied by the largest number of bands with inter-class DTW values greater than the intra-class distance (Table~\ref{tab:7}), indicating excellent crop discrimination capability. 

The visual comparisons and DTW-based analysis demonstrate that appropriate preprocessing methods enhance both class separability and the interpretability of spectral-temporal data. Accordingly, selecting optimal preprocessing strategies is critical to maximizing the accuracy and reliability of crop mapping models.

\subsection{Evaluation of supervised crop classification workflows} \label{sec:cropclassification}

\begin{figure}
	\centering
		\includegraphics[scale=0.44]{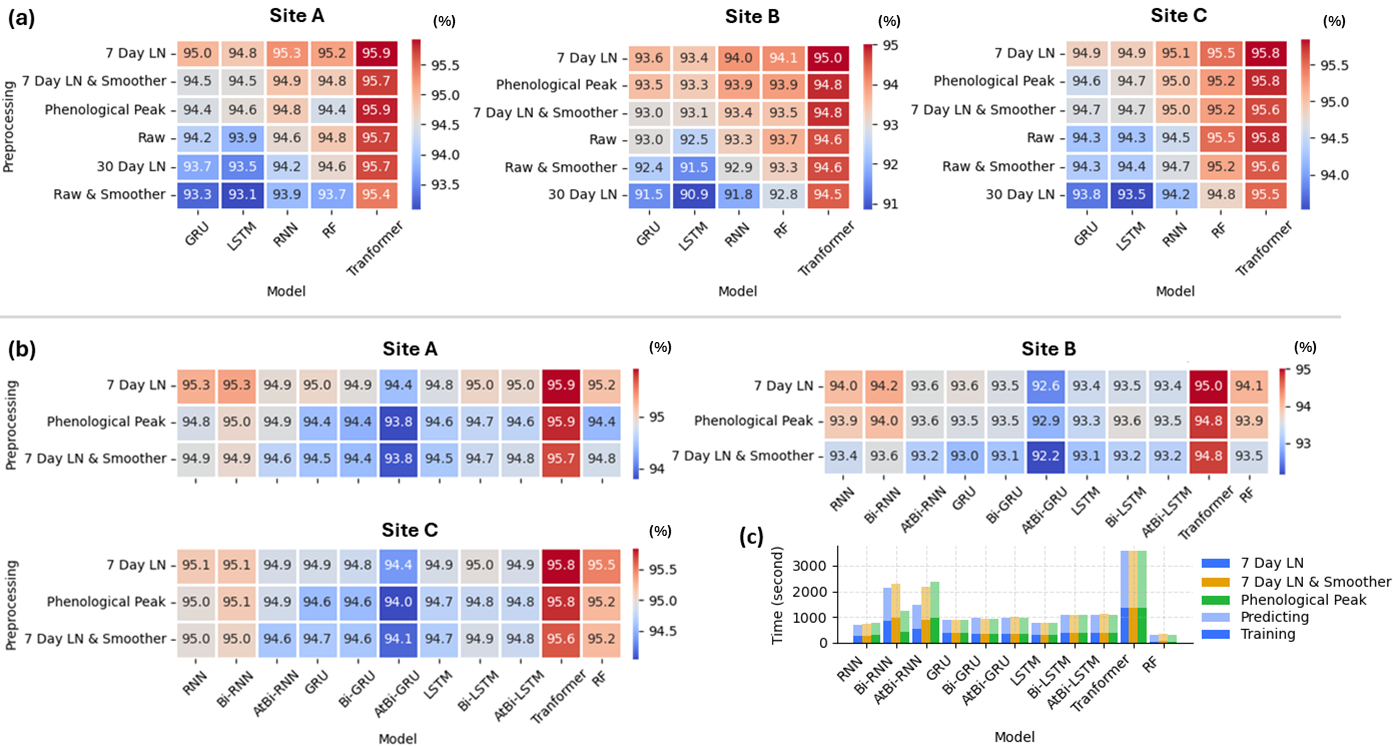}
    \caption{Classification accuracy and computational cost of six preprocessing methods and eleven pixel-wise classification models at three Corn-Soybean Belt sites. (a) Overall accuracy of six preprocessing methods combined with five representative models, sorted by ascending average accuracy. Three finer-scale linear resampling-based methods show consistent superiority. (b) Overall accuracy of eleven models paired with three optimal preprocessing methods, highlighting the Transformer's consistent top performance. (c) Training and full-scene prediction time for workflows at Site A shows that Transformer models require significantly longer times, while RF offered rapid training and prediction.}
	\label{fig:10}
\end{figure}

This section evaluated fully supervised crop classification workflows, following a two-tiered approach: first, we evaluated classification performance across six preprocessing approaches (Section~\ref{sec:preprocessing_result}); second, we compared eleven classification models based on the selected optimal preprocessing methods (Section~\ref{sec:model_results}). Sites A, B, and C span five degrees of latitude within the Corn and Soybean Belt and were characterized by stable crop rotations and relatively balanced class distributions, facilitating cross-site comparison. Data partitions in these supervised training sites (source sites) follow Table~\ref{tab:5}. Overall accuracy (OA) served as the primary evaluation metric, enabling concise, high-level comparisons across workflows. See Table~\ref{tab:6} for detailed model parameter configurations.

\subsubsection{Preprocessing method evaluation} \label{sec:preprocessing_result}
 
Building on the prior visual and DTW analysis of time series reconstruction methods (Section~\ref{sec:analysis_result}), we conducted additional classification experiments to assess the practical performance of six preprocessing approaches. These approaches were evaluated using five representative models: RF, Transformer, and three single-layer RNNs (RNN, LSTM, and GRU), with ten-fold cross-validation to identify the most effective preprocessing method for supervised crop classification (Fig.~\ref{fig:10}a). 

We observed that three preprocessing methods with 23 time steps--7-day linear resampling, phenological peak-focused resampling, and 7-day linear resampling with smoothing--consistently outperformed others. Specifically, 7-day linear resampling achieved the highest average OA on the test set: 95.2\% at Site A, 94.0\% at Site B, and 95.2\% at Site C. These results affirm that dense and temporally consistent spectral-temporal input is crucial for accurate crop classification \citep{Griffiths19}. The phenological peak method adapts to pixel-specific growing seasons, avoiding misalignment from fixed calendar windows, and making it adaptable for broader geographic applications. However, it slightly underperformed compared to fixed-period 7-day linear resampling, potentially because it struggles to capture meaningful peak signals for non-crop classes such as bare land and water body within the "other" category. The smoothed 7-day linear resampling consistently underperformed compared to the 7-day linear resampling as well, likely due to oversmoothing, which caused the loss of within-season spectral variability.

Interestingly, despite the presence of substantial noise and short time series (9, 7, and 9 time steps at Sites A-C, respectively), both RF and Transformer models trained on raw time-sequence data achieved competitive classification accuracy. For Transformers, this may stem from their self-attention mechanism, which can suppress noisy or uninformative observations \citep{MarcRußwurm20}. RF performance likely benefited from high spectral fidelity and the ability to leverage a large, diverse sample set (each site providing up to 400,000 trusted samples), which helped compensate for cloud-contaminated or missing observations. However, raw data limited the performance of single-layer RNNs, which are more sensitive to noise. The scalability of the raw-based method was also restrained by inconsistent temporal alignment across scenes.

\subsubsection{Model architecture evaluation} \label{sec:model_results}

We further evaluated eleven classification models using the top three preprocessing approaches (Fig.~\ref{fig:10}b). Model performance varied substantially across architectures. Transformer consistently achieved the highest classification accuracy across all sites (average OA: 95.8\% at Site A, 94.9\% at Site B, and 95.7\% at Site C). RF also performed competitively, particularly at Site C, outperforming several DL models despite not being designed for temporal data modeling. Within the RNN family, increasing architectural complexity (e.g., bidirectional or attention-enhanced variants) did not consistently lead to improved performance. In many cases, simpler architectures, such as RNN, Bi-RNN, GRU, or LSTM, achieved comparable or better accuracy, suggesting that they are sufficient for three-class crop mapping when numerous labeled samples and well-preprocessed time series are provided. The replicable results in Fig.~\ref{fig:10}a show that the difference in OA between the best- and worst-performing workflows ranged from 2.3\% to 4.1\%, indicating that workflow design, including preprocessing method and model selection, remains a crucial factor impacting large-scale crop mapping performance.

Training time and computational cost also varied significantly across workflows (Fig.~\ref{fig:10}c). Transformer models, while highly accurate, require considerably longer training durations and are best suited for GPU-enabled environments. In contrast, RF offered rapid training and lower total training-prediction time while maintaining strong classification performance. These characteristics make RF particularly appealing for applications based on CPU environments that require real-time analysis or large-scale inference. Overall, the findings support two practical supervised crop classification strategies. For high-accuracy applications with sufficient computational resources, Transformer models paired with 7-day linear resampling consistently achieved the highest accuracy. In contrast, RF models using the same inputs provided competitive performance with lower computational costs, making them suitable for resource-constrained applications. These results allow supervised workflows to be tailored based on the trade-offs between classification accuracy and operational constraints.

\subsubsection{Sample size and additional variables} \label{sec:samplesizevariables}

\begin{figure}
	\centering
		\includegraphics[scale=0.23]{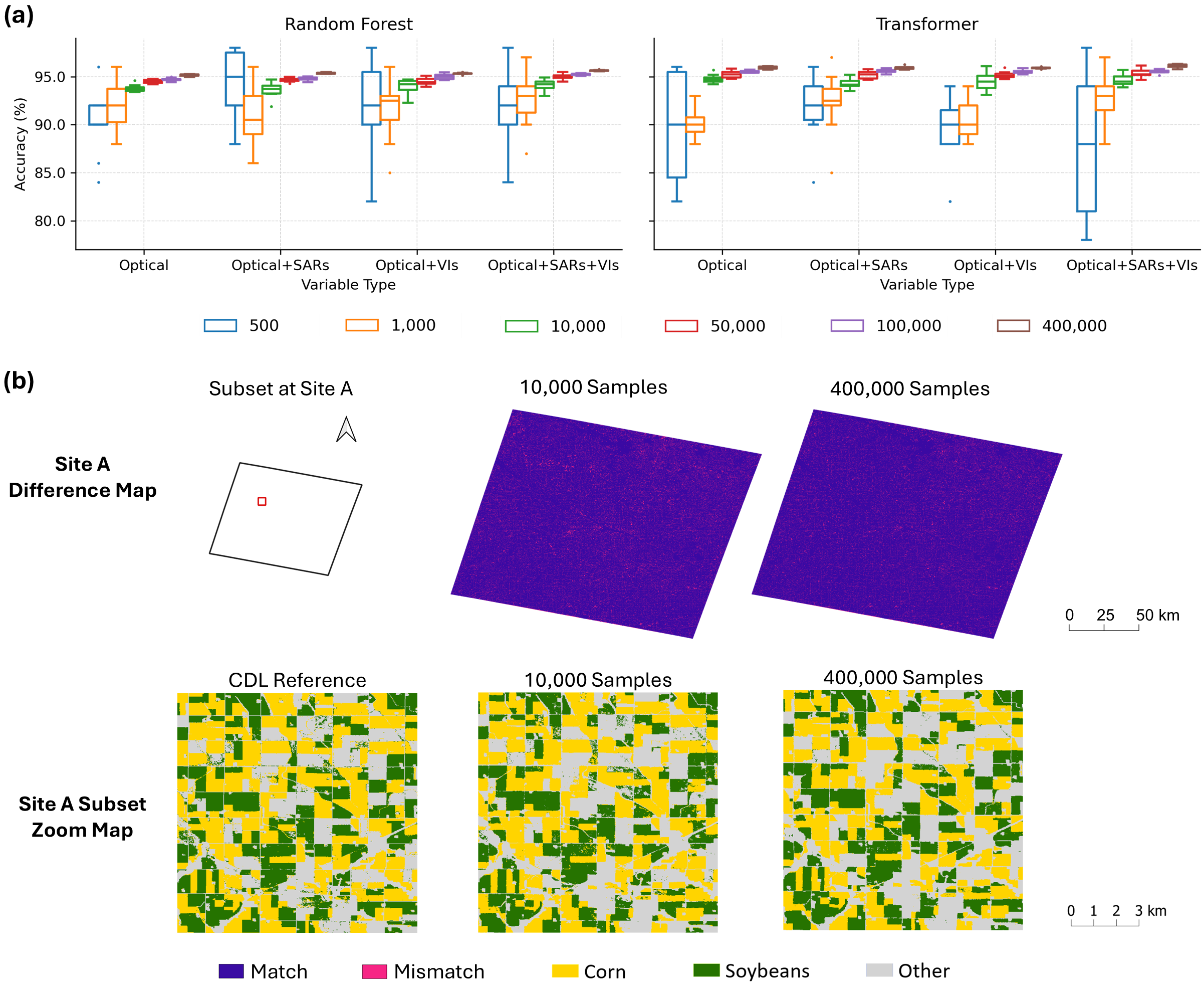}
    \caption{Six sample sizes and four variable combinations were evaluated on optimal RF and Transformer workflows. (a) Boxplot of 10-fold cross-validation overall accuracies. Accuracy improves and becomes more stable as the training sample size increases, though gains diminish at larger sample sizes. Incorporating all complementary variables enhances RF classification performance relative to using optical bands alone, while the Transformer exhibited relatively stable performance across variable sets. (b) Difference maps and prediction maps generated by the Transformer model using optical bands (subset area: \textasciitilde225 km$^2$ at Site A) illustrate that, while accuracy gains plateaued, increasing sample size continued to improve map quality. The difference map indicates the areas of mismatch between the prediction map and the CDL reference. Abbreviations: CDL, 2023 Cropland Data Layer reference map, with 85\% to 95\% producer’s accuracy \citep{cdl_acc}.}
	\label{fig:11}
\end{figure}

Unlike previous studies that examined either the effect of sample size \citep{FuYangyang23,waldner19} or feature selection alone \citep{ZhiFeng22}, this section explored the joint impacts of sample size and complementary variables on crop mapping performance at Site A in 2023. The focused experiments evaluated six-level sample sizes, ranging from small (500 and 1,000 samples) to very large (400,000 samples), alongside four combinations of input variables: optical spectral bands alone, optical bands combined with six VIs, optical bands combined with three SAR bands, and all optical, VIs, and SAR variables combined (Fig.~\ref{fig:11}; see Table~\ref{tab:5} for data partitioning details). The workflow consists of the best-performing model (RF or Transformer) and the optimal preprocessing method (7-day linear resampling). The VIs time series was fused with the spectral time series using input-level data fusion, and the SAR variable time series used feature-level data fusion to retain all information.

Overall accuracy from 10-fold cross-validation demonstrated a clear trend of increasing accuracy and more stable prediction outcomes with growing sample sizes, as evidenced by elevated median OA and reduced cross-validation interquartile range (Fig.~\ref{fig:11}a). While accuracy gains plateaued beyond 10,000 samples for both RF and Transformer models, large training datasets were still beneficial for generating high-quality pixel-wise prediction maps (Fig.~\ref{fig:11}b). Transformer-based predictions trained on excessive samples exhibited improved field homogeneity, and the 400,000-sample model achieved superior boundary delineation compared to the CDL reference. In practical terms, a sample size exceeding 3,000 for each class can be a sufficient threshold to produce reliable maps (OA over 90\%) on a typical Landsat scene (185 km × 170 km) in the Corn and Soybean Belt, suitable for decision-making with moderate post-processing.

The integration of additional variables (VIs and SAR variables) consistently improved classification accuracy across all sample sizes. RF models benefited the most, showing substantial gains when all optical, VIs, and SAR inputs were combined. Adding complementary variables improved the classification accuracy of RF beyond the next higher sample size level, meaning that the additional variables can compensate for sample limitations. In contrast, the Transformer trained with sufficient samples exhibited relatively stable performance across variable sets, with minimal changes in median OA and only slight increases in the range of cross-validation accuracies as variable complexity increased. These results highlight a key distinction: while both models improved with increased training data, RF more strongly leveraged multimodal inputs, suggesting that the value of additional variables may diminish for DL models when ample data are already available.

\subsection{Evaluation of workflow direct transferability across domains} \label{sec:directtransferability}

To assess the temporal, sensor, and spatial transferability of classification workflows, we compared models pre-trained on source-domain data (as described in Section~\ref{sec:model_results}) with upper-bound models trained within each target site and year. For pre-trained models, we fixed the input configuration to optical bands only and used the maximum available training sample size (see Table~\ref{tab:5}), based on earlier findings that larger sample sizes consistently improved performance, while adding complementary variables yielded limited gains for DL models (Section~\ref{sec:samplesizevariables}). Upper-bound models were trained using 9,000 labeled trusted samples from the target domain using 9-fold cross-validation (8,000 for training and 1,000 for validation). This sample size was chosen to ensure relatively stable and competitive model performance (as described in Section~\ref{sec:samplesizevariables}) and match the maximum available labeled samples at Site E (a spatial transfer target site), thereby supporting consistent comparisons across domains. All models were evaluated on the kept aside, fixed 4,000-sample test set at the corresponding target site, allowing direct performance comparison under identical test conditions (Fig.~\ref{fig:12}).

\subsubsection{Temporal transfer} \label{sec:temporalDirectTransfer}
\begin{figure}
	\centering
		\includegraphics[scale=0.58]{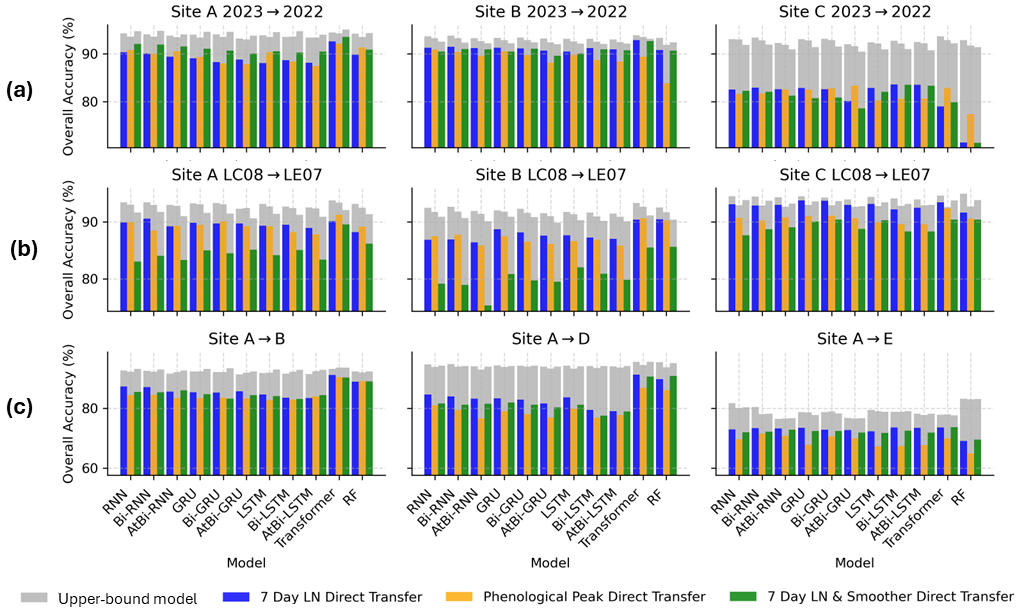}
    \caption{Comparison of model transferability across temporal, sensor, and spatial domains. Each grouped bar plot summarizes the overall accuracy of eleven pixel-wise classification models combined with three optimal preprocessing methods. Colored bars represent the OA of transferred pre-trained models, while gray bars indicate the upper-bound accuracies achievable in the target domain via supervised learning, assuming substantial target-domain data. The subplots represent (a) temporal transfer (from 2023 to 2022), (b) sensor transfer (Landsat 8 OLI to Landsat 7 ETM+ in 2022), and (c) spatial transfer within the year 2023. The average overall accuracy was calculated from 8 out of 10 cross-validation folds, excluding the lowest and highest fold results to minimize the influence of extreme cases.}
	\label{fig:12}
\end{figure}

Temporal transfer experiments examined the robustness of models trained on 2023 LC08 data when applied to 2022 LC08 in the same area (Fig.~\ref{fig:12}a). Three transfer pairs at Sites A, B, and C were selected due to their consistent cropping systems and high-quality trusted labels in both years, enabling reliable year-to-year evaluation. In all direct transfer pairs, the upper-bound workflow yielded higher accuracy than the direct transfer workflow, underscoring the value of temporally relevant training data. 

Site C presented the greatest challenge for temporal transfer, primarily due to marked phenological and environmental discrepancies between 2022 and 2023. In 2022, cooler and wetter early-season conditions followed by intensified late-season drought resulted in greater hydrological variability and a growing season-wide delay in NDVI dynamics compared to 2023 (Appendix~\ref{sec:PhenologyAnalysis}; \citet{ncei2022december}). Under these challenges, all transferred models exhibited reduced performance, but DL architectures were generally more resistant than RF. RF paired with phenological peak-based preprocessing demonstrated greater resistance to interannual environmental variability than when using fixed-period methods. On the contrary, when interannual phenology is relatively consistent, evidenced by similar Moisture Stress Index (MSI; \citet{msiS2}) and NDVI trajectories across years at Sites A and B (Appendix~\ref{sec:PhenologyAnalysis}), directly transferred workflows remained highly effective. Transformer models with fixed-period preprocessing achieved the highest direct-transfer accuracies, closely approaching the performance of upper-bound models. 

These results highlight the sensitivity of temporal transfer to phenological variation. At sites where crop growth and weather conditions remained consistent across years, fixed-period preprocessing paired with any model, whether ML or DL, achieved high transfer accuracy. In contrast, under substantial phenological and environmental variability between years, DL models showed less performance degradation than ML.

\subsubsection{Sensor transfer}

Sensor transfer evaluations assessed model performance when transferring from OLI (LC08) to ETM+ (LE07) imagery collected in the same area and year (Fig.~\ref{fig:12}b). In 2022, Sites A, B, and C had full year-round temporal coverage from both sensors, enabling consistent cross-sensor comparisons using identical sample locations. Across all sites, upper-bound models trained and tested on LE07 consistently outperformed directly transferred models trained on LC08, indicating notable spectral mismatches between the two sensors. 

Preprocessing methods had a greater influence on direct-transfer accuracy than the model architecture. The 7-day linear resampling method achieved the highest direct-transfer accuracy at all three sites (average OA: 89.6\% at Site A, 87.9\% at Sites B, and 92.9\% at Site C), followed by the phenological peak-based method (average OA: 89.3\% at Site A, 87.4\% at Site B, and 90.6\% at Site C). In contrast, the smoothed 7-day linear resampling consistently underperformed. The reduced accuracy of the smoothed 7-day linear resampling method likely stems from the WE smoother's limited ability to reconstruct ETM+ spectral trajectories that align with OLI, particularly in regions where LE07 gap-filled data introduce inconsistencies. The phenological peak-based method also showed unstable performance, potentially due to discrepancies in NDVI peak timing and extraction windows between data from the two sensors. These findings highlight that both the wavelength differences between the two Landsat sensors and structural inconsistencies of Landsat 7, including the scan-line corrector failure in ETM+, introduce challenges for direct model transfer. Robust preprocessing strategies that harmonize multi-sensor time series and mitigate sensor-specific noise are essential for accurate cross-sensor classification.

\subsubsection{Spatial transfer} \label{sec:spatialDirectTransfer}

Spatial transfer experiments assessed the generalizability of models trained at source Site A when applied to geographically distinct target Sites B, D, and E in the same year, using the same data product (Fig.~\ref{fig:12}c). Site A was selected as the source domain due to its substantial number of trusted samples, providing a robust foundation for pre-trained models. Three transfer pairs varied in domain shifts regarding cropping systems, latitude, and climate zone, allowing for spatial transfer assessment from similar (A to B) to more challenging scenarios (A to D, A to E). Across all transfer scenarios, upper-bound models consistently outperformed directly transferred workflows, underscoring the importance of locally representative training data for robust classification performance.

Sites A and B, both situated within the Corn-Soybean Belt, demonstrated consistently high direct-transfer accuracies. In contrast, classification performance declined when transferring models from Site A to climatically distinct Site D, which lies in a different climate zone (Cfa) and the soybean-dominant belt. Performance further deteriorated at geographically distant Site E, which is located on another continent, a different climate zone (Cfa), and characterized by smaller farms and more heterogeneous land use. 

In addition to the domain-shift effect, model architecture also affects transferability. Workflows that consist of Transformer or RF models combined with fixed-period preprocessing demonstrated greater resistance to small to moderate environmental changes than the other combinations and approached the performance of the upper-bound models. All findings highlight that effective spatial transfer strongly depends on workflow design and domain similarity.

\subsubsection{Multi-source domain direct transfer}\label{sec:crosstransfer}

\begin{figure}
	\centering
		\includegraphics[scale=0.52]{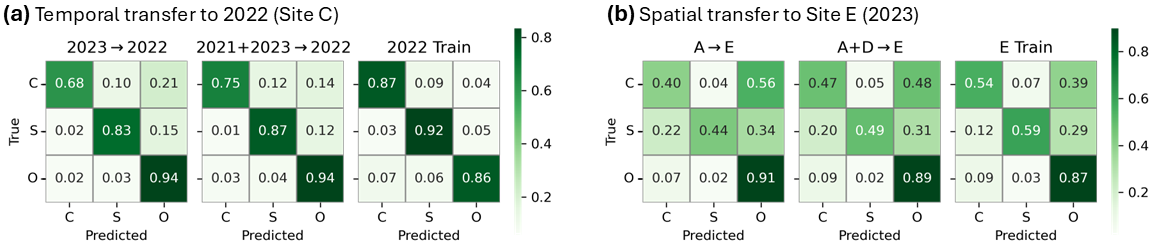}
    \caption{Confusion matrices for target-site test sets under (a) temporal transfer at Site C and (b) spatial transfer to Site E, comparing single-source, multi-source, and fully supervised models. Fully supervised Transformer models were trained on 9,000 labeled samples from each target site. Class labels are corn (C), soybean (S), and "other" (O). Values represent row-normalized proportions; higher diagonal values reflect stronger per-class producer's accuracy. Multi-year training improved temporal transfer performance relative to single-year training, while combining data from multiple source sites enhanced crop classification accuracy but did not improve "other" class performance.}
	\label{fig:13}
\end{figure}

To extend the evaluation beyond single-source domain transfers (Sections~\ref{sec:temporalDirectTransfer} -~\ref{sec:spatialDirectTransfer}), we investigated whether composite-domain data could improve model generalizability under challenging temporal and spatial transfer scenarios. This experiment reflects many transfer practices in large-scale crop classification, such as utilizing historical (composite-year) or multi-site (composite-site) samples to improve model generalizability \citep{WijesinghaJayan24, ZhangShibo24, LiuZhe22, LinZhixian22}. Experiments based on two of the most challenging scenarios: temporal transfer from 2023 to 2022 at Site C and spatial transfer from Site A to Site E (Fig.~\ref{fig:13}). The workflow comprised 7-day linear resampling, multi-source domain sample generation, and the Transformer model, which examined optimal in single-source domain transfer scenarios. Multi-source domain training samples were evenly collected from all domains to hold constant total training sample sizes across settings.

For temporal transfer, integrating data from 2021 and 2023 improved classification accuracy across all classes compared to the model trained on 2023 samples alone. Incorporating environmental and phenological diversity in 2021 enhanced the model's generalizability to the 2022 target samples, demonstrating the effectiveness of multi-year training. In contrast, spatial transfer results showed more limited benefits from multi-site training. Although adding samples from Site D (same climate zone as Site E; diverse cropping systems and land use) improved corn and soybean accuracy, it failed to enhance classification for the heterogeneous "other" class. 

These findings indicate that direct transfer has potential for both single-source and multi-source domain workflows in the case of low domain shift. The multi-year model further reduces the magnitude of domain shift in temporal transfer. However, direct transfer remains insufficient under more significant domain shifts. Therefore, incorporating adaptive strategies into transfer learning workflows is crucial, especially in moderate to large domain shift scenarios.

\subsection{Evaluate adaptive transfer learning approaches across spatial domains} \label{sec:improvetransfer}

In this section, we examined UDA and fine-tuning approaches in spatial transfer scenarios ranging from regionally similar (A to B) to highly divergent (A to E). The UDA approach, based on a DANN, was trained with 400,000 labeled source-site samples and the spectral-temporal features of 4,000 unlabeled samples from each target site (Fig.~\ref{fig:14}). In contrast, the fine-tuning approach involved updating all parameters of the pre-trained model using 4,000 labeled samples from each target site (see Table~\ref{tab:5} for data partitions). For the most challenging transfer case (A to E), we further evaluated advanced fine-tuning strategies designed to address class imbalance in the target domain. Due to sample class imbalance in Sites D and E, we evaluated the performance of both approaches primarily by per-class producer's accuracy (PA).

\subsubsection{Unsupervised domain adaptation} \label{sec:domainAdaptationTechniques}

\begin{figure}
	\centering
		\includegraphics[scale=0.55]{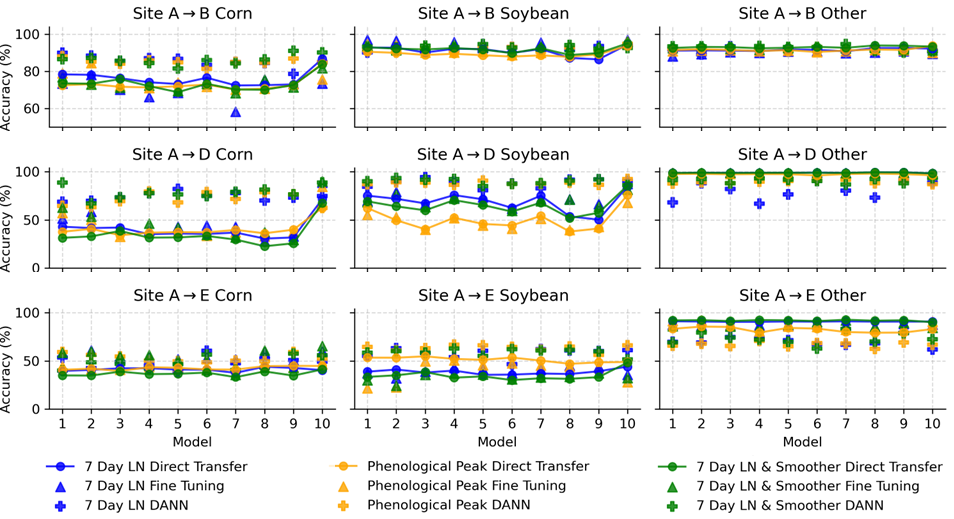}
    \caption{Comparison of per-class producer's accuracy for direct transfer, DANN, and complete fine-tuning workflows across spatial transfer scenarios. Rows represent spatial transfer pairs with varying levels of domain shifts (A to B: low, A to D: moderate, A to E: challenging.) Within each panel, DL models are ordered left-to-right: 1) RNN, 2) Bi-RNN, 3) AtBi-RNN, 4) GRU, 5) Bi-GRU, 6) AtBi-GRU, 7) LSTM, 8) Bi-LSTM, 9) AtBi-LSTM, and 10) Transformer. Lines with circular markers represent the direct-transfer baseline for each of the three preprocessing methods. Adapted models (DANN and fine-tuning) are shown as filled plus and upward-pointing triangle symbols, respectively. Across transfer scenarios, DANN significantly improved accuracy for corn and soybean relative to direct transfer, but generally struggled with the heterogeneous "other" class. Fine-tuning provided slight yet consistent accuracy improvements across all classes, with infrequent overfitting. Average accuracy calculated from 8 of 10 cross-validation folds, omitting the lowest and highest to reduce the impact of outliers.}
	\label{fig:14}
\end{figure}

DANN significantly improved the classification accuracy of target crops for nearly all DL workflows (Fig.~\ref{fig:14}). This is reflected in corn and soybean classification results (filled-plus symbols) consistently meeting or exceeding the corresponding direct transfer results (lines). For the relatively similar domain pair (A to B), DANN not only improved target crop classification but also maintained performance on the heterogeneous "other" class, resulting in higher OA (Appendix~\ref{sec:transferlearningOA}). In the moderate domain shift scenario (A to D), direct transfer yielded unreliable results for corn (often with accuracy below 50\%), whereas DANN adaptation substantially recovered performance for both corn and soybean. However, this gain came with the reduced "other" class accuracy, likely due to its greater heterogeneity and domain-specific variability. In the most challenging transfer case (A to E), DANN continued to improve target crop accuracy, but the resulting performance remained unreliable, and "other" class accuracy dropped sharply, indicating that the current UDA configuration does not fully address extreme domain shifts.

\begin{figure}
	\centering
		\includegraphics[scale=0.23]{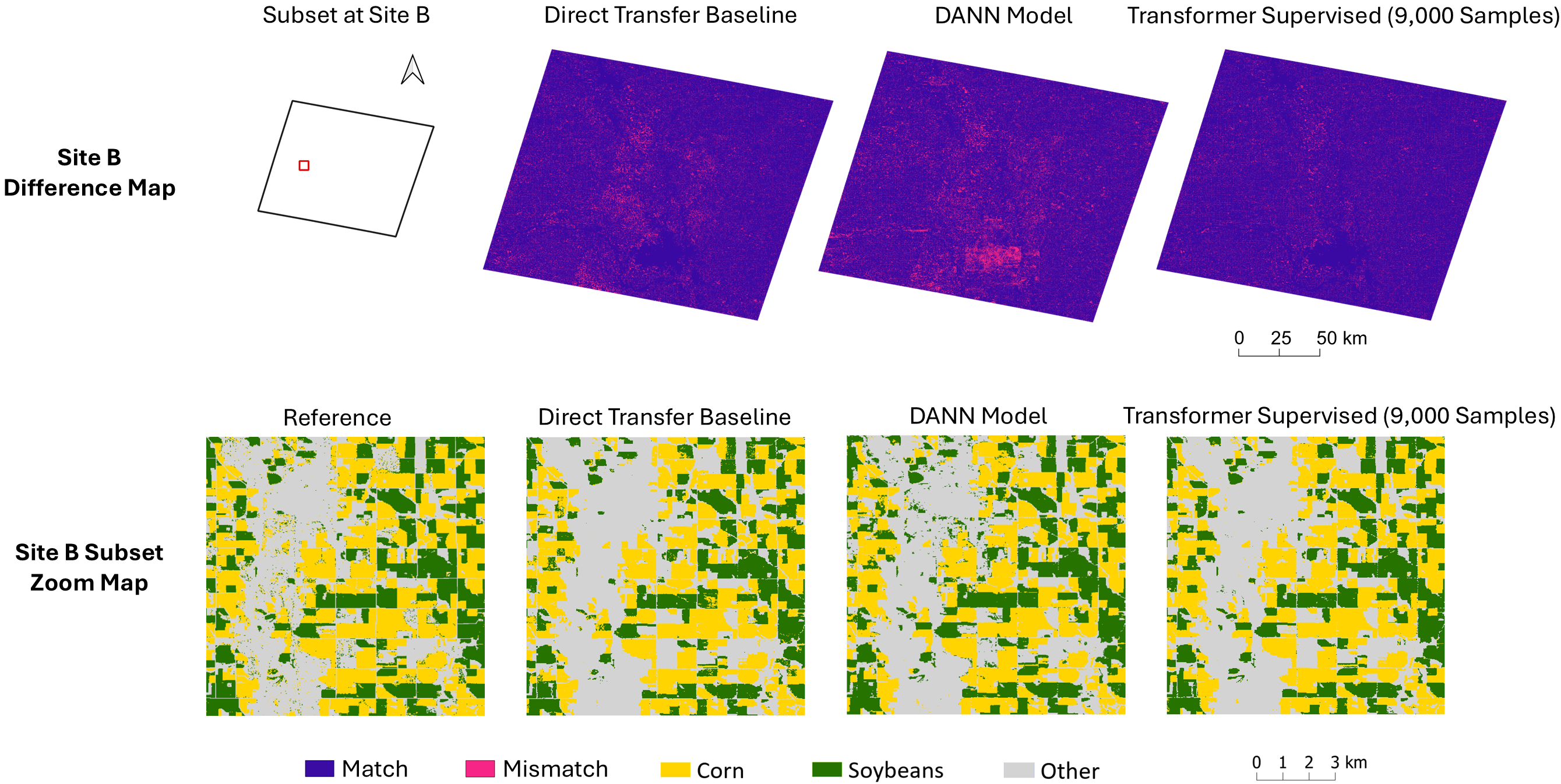}
    \caption{Difference maps and prediction maps at Site B, generated using direct transfer, DANN-based transfer learning, and fully supervised learning workflows (zoomed area covers \textasciitilde225 km$^2$). Under low domain shift, both direct transfer and DANN-based models can achieve high-accuracy crop mapping without requiring labeled data from the target site (over 90\% OA). The reference map was CDL in 2023, with 85\% to 95\% producer’s accuracy \citep{cdl_acc}. The difference map exhibits the areas of mismatch between the prediction map and the CDL reference. Abbreviation: DANN, Domain-Adversarial Neural Networks.}
	\label{fig:15}
\end{figure}

The DANN approach's effectiveness was further evaluated by comparing its predictions with direct transfer, CDL reference map, and a fully supervised model at Site B (Fig.~\ref{fig:15}). All transfer models leveraged knowledge from the source site (Site A) without using labeled data from Site B. The supervised Transformer model was trained with 9,000 trusted samples from Site B as a benchmark. Among the direct transfer results, the Transformer with 7-day linear resampling achieved the highest OA (92.1\%). For DANN models, the AtBi-LSTM architecture with smoothed 7-day linear resampling yielded the best OA (91.2\%), representing a substantial improvement over its direct transfer performance (84.9\%). The DANN approach showed significant misclassification in non-agricultural areas, particularly in Omaha, NE, due to land use differences between corn and soybean-dominated Site A and urban-heavy Site B. This misclassification could be reduced through additional post-processing. The fully supervised model achieved the highest OA (95.2\%), surpassing all transfer-based methods. Nevertheless, all three approaches, direct transfer, DANN, and full supervision, produced strong generalization at Site B with OA exceeding 90\%. These results suggest that, under low domain shift, both direct transfer and DANN-based models can achieve high-accuracy crop mapping without requiring labeled data from the target site.

\subsubsection{Advanced fine-tuning strategy}

Compared to DANN, complete fine-tuning produced smaller yet more consistent improvements across all three classes (Fig.~\ref{fig:14}). The evaluated fine-tuned model corresponds to its early-stopped checkpoint selected based on minimum validation loss. While it didn't match DANN's improvements in corn and soybean classification under low to moderate domain shift scenarios, it consistently provided strong classification results for the "other" class. Notably, for the most challenging transfer (A to E), fine-tuning constantly improved OA, whereas DANN led to a net decline in performance (Appendix~\ref{sec:transferlearningOA}). This highlights the robustness of fine-tuning, especially when limited but labeled target-domain data is available. Fine-tuning offered a slight but reliable improvement in accuracy compared to direct transfer and may be preferable when domain gaps are substantial or when a balanced performance increase across all classes is required.

\begin{table}[ht]
  \centering
  \caption{Fine-tuning strategies and training configurations}
  \footnotesize
  \begin{tabular}{>{\centering\arraybackslash}p{1.5cm} p{9.5cm} >{\centering\arraybackslash}p{1.5cm}}
    \hline
    \textbf{Strategy} & \textbf{Fine-tuning Design} & \textbf{Epoch}\\
    \hline
    R1 & Complete fine-tuning using the full dataset with a standard shuffled dataloader. & 40 \\ \\
    \hline
    R2 & Fine-tuning with class-weighted loss and weighted random sampler based on inverse class frequency. & 40 \\
    \hline
    R3 & Fine-tuning on a class-balanced subset created via undersampling to match minority class size. & 40  \\
    \hline
    R4 & Two-stage fine-tuning: (1) the same as R3, (2) continue with R2 with all layers frozen except the output head. & 20/stage \\
    \hline
    \end{tabular}
  \label{tab:8}
\end{table}

\begin{figure}
	\centering
		\includegraphics[scale=0.6]{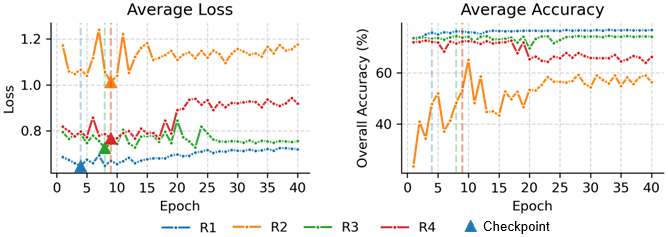}
	\caption{Training loss and validation accuracy curves for four fine-tuning strategies, including basic complete fine-tuning (R1), class-weighted sampling and loss (R2), balanced subset undersampling (R3), and two-stage fine-tuning (R4). The pre-trained models used training samples from Sites A and D.}
	\label{fig:16}
\end{figure}

Considering the imbalanced sample class distribution at Site E, as indicated by the high "other" class accuracy and low target-crop accuracies, we explored three advanced fine-tuning strategies to address this issue, aiming to obtain better generalizable and adapted models (Table~\ref{tab:8}). Four fine-tuning approaches were examined based on an optimal workflow consisting of 7-day linear resampling preprocessing and a pre-trained Transformer, including basic complete fine-tuning (R1), class-weighted sampling and loss (R2), balanced subset undersampling (R3), and two-stage fine-tuning (R4). Training loss and validation accuracy curves were used to evaluate model adaptation (Fig.~\ref{fig:16}).

\begin{figure}
	\centering
		\includegraphics[scale=0.52]{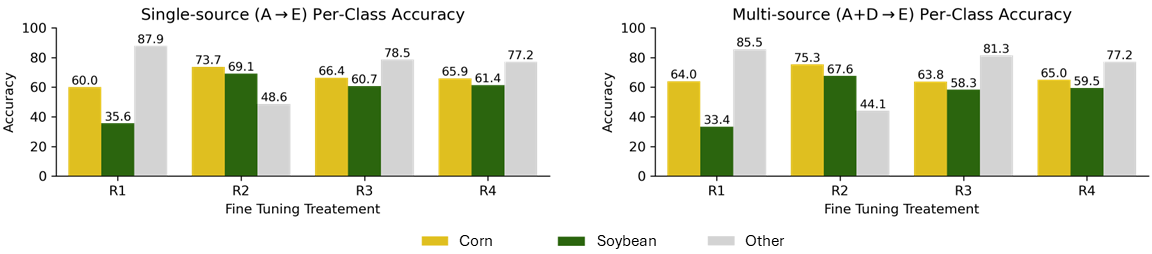}
    \caption{Per-class producer's accuracy from single-source and multi-source domain fine-tuning checkpoints. Multi-source domain fine-tuning did not yield clear advantages over single-source approaches under large domain shifts, consistent with the findings in direct transfer results. The average accuracy was calculated from 8 out of 10 cross-validation folds, excluding the lowest and highest fold results to minimize the influence of extreme cases.}
	\label{fig:17}
\end{figure}

Basic complete fine-tuning (R1) initially showed low training loss and high accuracy but rapidly overfitted the dominant class ("other"), resulting in poor generalization to the minority soybean class (Fig.~\ref{fig:17}). Class-weighted sampling and loss (R2), intended to mitigate class imbalance, resulted in persistently high training losses and lower OA. While improving the detection of minority classes, it significantly increased misclassification rates for the "other" class, demonstrating the drawbacks of overly aggressive weighting schemes. Balanced subset undersampling (R3) proved to be the most effective strategy, stabilizing the training loss after approximately 25 epochs. It notably enhanced accuracy for target crop classes while maintaining decent performance on the "other" class. The two-stage fine-tuning (R4), sharing the first 20 epochs with R3, initially achieved stable training but subsequently showed increased instability and accuracy degradation, primarily misclassifying the "other" class as target crops. Consistent with findings from direct transfer experiments, fine-tuning with composite data from Sites A and D did not yield substantial improvements over single-source (Site A) transfer.

\begin{figure}
	\centering
		\includegraphics[scale=0.23]{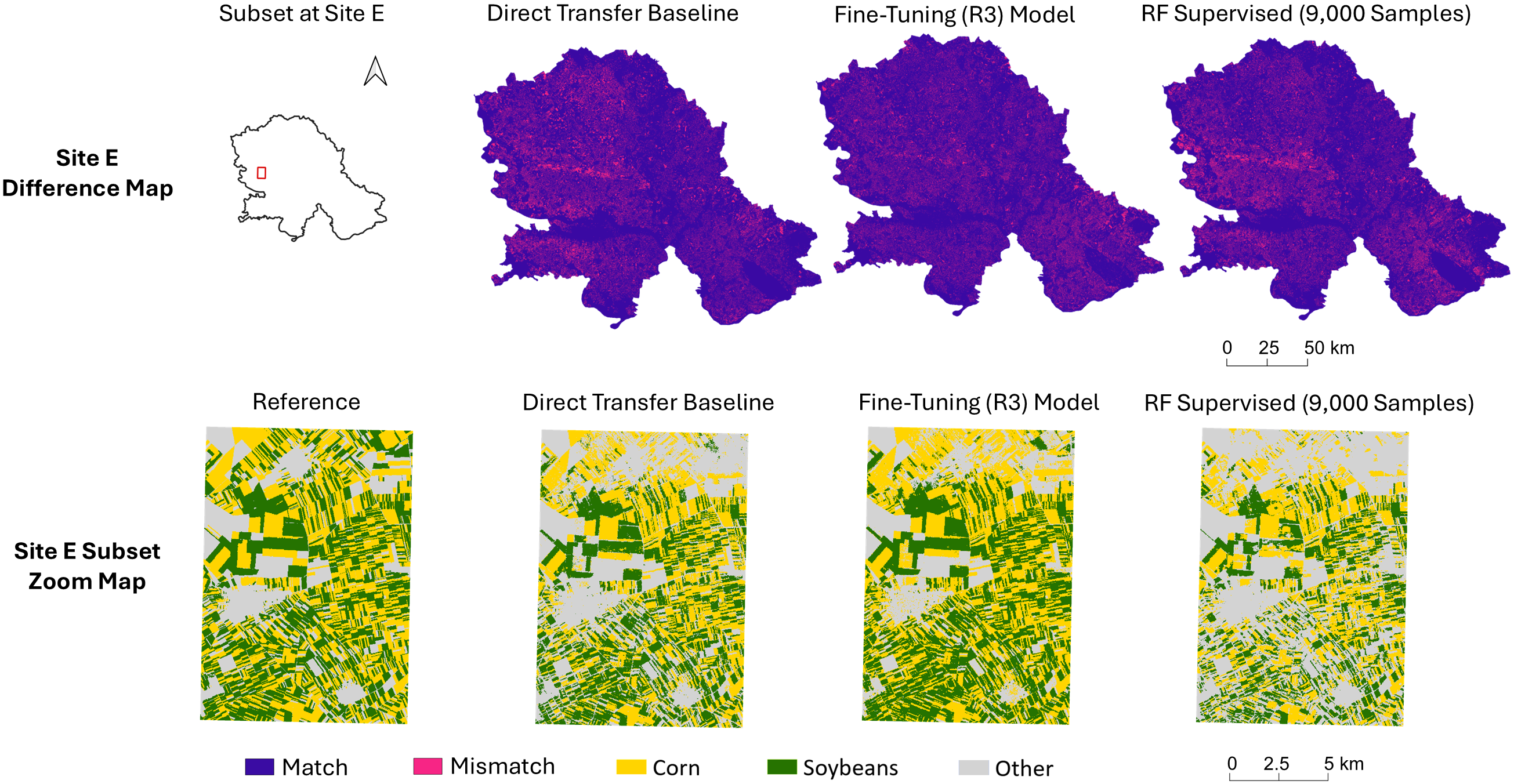}
    \caption{Difference maps and prediction maps at Site E, generated using direct transfer, fine-tuning R3-based transfer learning, and fully supervised learning workflows (zoomed area covers \textasciitilde225 km$^2$). The reference map was obtained from the BioSense Institute with a 10-m spatial resolution and an overall accuracy over 90\% \citep{MilosPandzic24, ZivaljevicBranislav24}. The difference map shows the areas of mismatch between the predicted map and the 30-m reference resampled using the nearest neighbor method. The fully supervised training RF exhibits a largely underestimated cropland area and significant fragmentation, while the fine-tuned R3 showed stronger generalization, particularly in larger field parcels.}
	\label{fig:18}
\end{figure}

Prediction map evaluations further validated the R3 strategy by assessing model generalizability over a broader area beyond sparse test points (Fig.~\ref{fig:18}). At Site E, we compared R3 to a direct transfer baseline, a reference map, and a fully supervised model. Each workflow used for prediction was selected based on the highest OA within its group, all employing 7-day linear resampling. The single-source direct transferred Transformer model and the fine-tuned R3 achieved OAs of 74.8\% and 76.1\%, respectively. The fully supervised RF model, trained on 9,000 labeled samples from Site E, reached 85.5\% OA. Despite its higher test accuracy, the supervised RF prediction exhibited low generalizability, according to a largely underestimated cropland area and significant fragmentation in the prediction map. In contrast, the fine-tuned R3 showed stronger generalization, particularly in larger field parcels, and outperformed direct transfer by leveraging knowledge from the balanced undersampling of 4,000 Site E samples. However, the fine-tuned R3 showed limited accuracy in smaller parcels, likely due to the 30-m Landsat 8 introducing mixed spectral signals in heterogeneous fields. This suggests that higher-resolution imagery, such as Sentinel-2 at 10 m, may be more suitable for improving transfer performance in smallholder farming regions.

\section{Discussion}

Grounded in the literature review and systematic workflow comparisons, this section provides practical insights in selecting preprocessing methods, model architectures, and transfer learning approaches for the large-scale pixel-wise crop mapping workflows for data-rich (Section~\ref{sec:discussion_supervised}) and data-scarce (Section~\ref{sec:discussion_adaptive}) areas.

\subsection{Optimal supervised and transferable crop mapping workflow} \label{sec:discussion_supervised}

Contrary to the widespread use of monthly resampling or aggregation in pixel-wise crop classification \citep{Xuan23, rusvnak2023}, this study demonstrated that such coarse (30-day) temporal aggregation often underperformed compared to both raw and finer-scale (7-day) methods. Confirmed previous studies indicating that shorter interval composites enhance classification performance by better capturing critical temporal information \citep{Griffiths19}. Our findings further revealed that monthly resampling could obscure important temporal changes, negatively impacting model accuracy, particularly for non-Transformer models. The 7-day linear resampling method proved to be the most robust preprocessing technique, consistently delivering strong performance in both supervised and transfer learning workflows, aligning with trends in recent crop classification research \citep{SinaMohammadi24, Xu20a}. Additionally, other fixed-period finer-scale methods, such as smoothed 7-day linear resampling, performed well in supervised crop mapping and remained robust in spatial and temporal transfer scenarios, though their effectiveness was limited in cross-sensor (OLI to ETM+) transfer.

The Transformer consistently delivered the highest classification accuracy across diverse preprocessing approaches, underscoring the effectiveness of the self-attention mechanism in capturing global spectral-temporal dependencies critical for crop mapping. This finding aligns with prior research suggesting that self-attention not only enhances the use of temporal information but also suppresses cloud-related noise \citep{Perich23, TangPengfei24} and produces cleaner, spatially contiguous prediction maps with reduced speckle and fragmentation \citep{AleissaeeAbdulazizAmer23}. This study further suggests that Transformers are robust to input variations such as irregular date intervals and coarse temporal resolutions (Fig.~\ref{fig:10}), making them well-suited for both pixel-wise and patch-based crop mapping, the latter of which often involves sparse image time series and residual cloud noise input \citep{FanLingling24}. Additionally, pre-trained Transformer models demonstrated strong transferability under low and moderate domain shifts, particularly when combined with fine-scale, fixed-period preprocessing. Notably, RF models, leveraging ensemble techniques, exhibited exceptional performance with finer-scale preprocessing or raw data under conditions of abundant training samples. Although the RF is more susceptible to the pepper effect than DL models, its rapid training and prediction capabilities make it particularly suitable for environments with limited computational resources. RF also shows strong generalizability at low domain shifts, whereas its performance degrades significantly compared to DL models in larger domain shifts.

\subsection{Adaptive workflows for pixel-wise large-scale crop mapping: supervised training versus transfer learning} \label{sec:discussion_adaptive}

\begin{figure}
	\centering
		\includegraphics[scale=0.42]{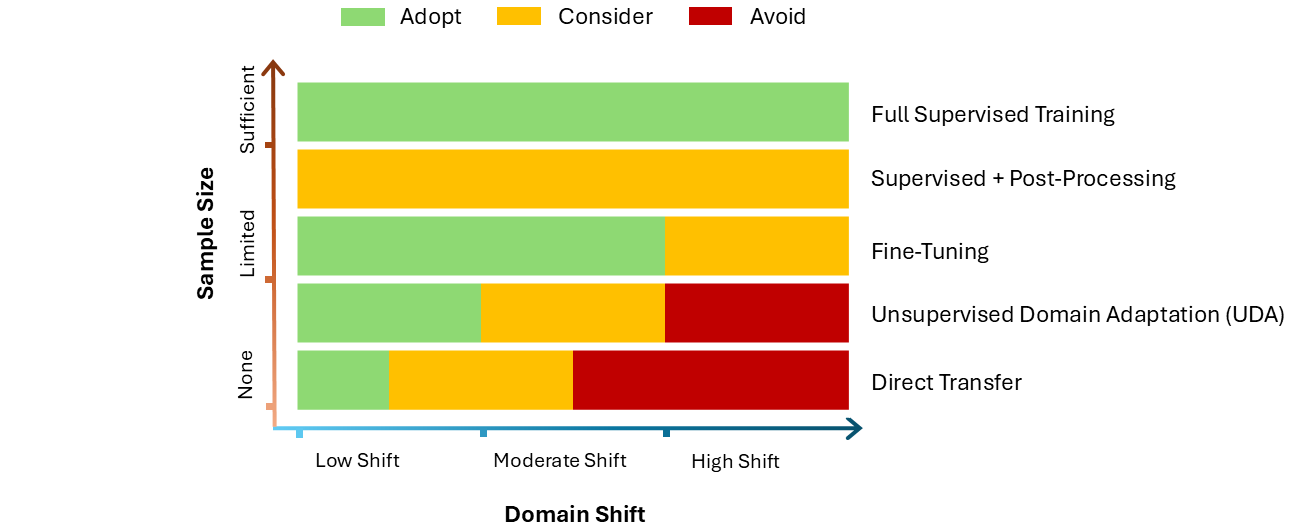}
    \caption{Workflow choice depends primarily on labeled sample availability, while transfer learning strategy selection is further influenced by the magnitude of domain shift. The sample size reflects the availability of labeled data at the target site, where the threshold between limited and sufficient data varies by spatial extent, crop types, and class balance. Domain shift reflects the degree of differences in landscape structure, phenological patterns, and environmental conditions between source and target domains. Green, yellow, and red segments indicate recommended, conditionally considered, and discouraged strategies, respectively. Conditionally considered strategies indicate that the workflow outperforms direct transfer while only achieving acceptable classification accuracy (typically 75–85\% OA in this study), which may be sufficient depending on the application. These flexible approaches support both high-accuracy crop mapping in well-surveyed regions and scalable mapping in underrepresented areas (if appropriate source domains exist), enabling more inclusive and data-driven agricultural monitoring.}
	\label{fig:19}
\end{figure}

We examined that well-trained RF, RNN-based, and Transformer supervised workflows were generally effective for direct transfer when the target site shares similar characteristics with the source (Fig.~\ref{fig:15}), but their performance declined with substantial differences in climate, field structure, phenology, or mixed-pixel conditions \citep{WijesinghaJayan24, Xu20a, ChenYaoliang16}. Multi-source domain approaches, which leverage composite training data from related source domains, proved effective in temporal transfer. Training the Transformer on multi-year samples effectively captures the phenomenon of phenological variability and reduces domain shifts, improving cross-year validation and temporal transferability, consistent with prior crop mapping practices \citep{WijesinghaJayan24, Xu20a}. The representative UDA method, the DANN transfer learning workflow, boosted classification performance for homogeneous target crop classes without labeled target data by encouraging models to learn domain-invariant features, thus enhancing adaptability under low to moderate domain shifts. However, DANN struggled with the heterogeneous class, limiting its effectiveness for highly divergent transfer scenarios. Previous studies on UDA in crop mapping have also shown success (acceptable test F1 score or OA higher than 70\%) mainly under moderate or low domain shifts \citep{LeiLei24, YumiaoW23, NyborgJoachim22}. Consequently, UDA is recommended primarily for temporal and spatial transfers with minor to moderate domain differences.

Fine-tuning consistently succeeds in adapting pre-trained models to diverse transfer scenarios \citep{MilosPandzic24, ArturN21}, and even to address large domain shifts \citep{YangLingbo23}. Advanced fine-tuning techniques, such as balanced subset undersampling, enhanced minority class accuracy while preserving acceptable majority class performance, demonstrating robustness with highly imbalanced target-site samples (Fig.~\ref{fig:18}). However, under low domain shifts, DANN outperformed fine-tuning and direct transfer by achieving higher crop-class accuracy without significant degradation of the "other" class. This indicates that DANN may be preferable for transfer scenarios involving abundant source-domain data and low domain shift. In this case, hybridizing DANN and fine-tuning, or using target-site labeled samples to post-process prediction maps, also holds significant promise.

Fully supervised training is reliable and generalizable only when the sample size exceeds a critical threshold (for instance, 3,000 samples for each class at Site B). \citet{NyborgJoachim22} suggested that fully supervised learning models of target sites consistently outperformed UDA models, \citet{MilosPandzic24} also proposed that once the sample size of target sites exceeds a certain limit, conventional training from scratch surpassed the fine-tuning approach. If the sample size continues to increase, the accuracy may reach a plateau, while exceedingly large sample sizes continue to contribute to improving model stability and generalizability. In addition, integrating complementary variables, including VIs and SAR variables, notably enhanced RF classification accuracy, compensating for sample-size limitations. However, such benefits were marginal for Transformer models.

Currently, the majority of operational crop mapping relies on fully supervised training with ample labeled data. However, this review highlights the flexible, dynamic choice of supervised training versus transfer learning, as determined by the labeled sample size and the presence of appropriate source domains (Fig.~\ref{fig:19}). Ultimately, systematically evaluating and adapting these workflows helps achieve two complementary objectives: producing high-quality crop maps in regions with extensive survey data, and expanding the generation of reliable crop maps in underrepresented or less-surveyed areas and years. This dual approach fosters innovations in large-scale agricultural monitoring, yield estimation, and potentially data-driven decision making to support global food security.

\section{Future work}

While this study provides a comprehensive evaluation of pixel-wise classification workflows, several promising directions remain unexplored and could substantially enhance crop mapping practices. First, this study focused exclusively on single-season crop classification. Future research could extend this analysis to multi-season crops and diverse crop types, enabling workflows to generalize across broader regions. Second, while this work standardized hyperparameter settings across models to facilitate a fair horizontal comparison of architecture performance, it did not explore model-specific or site-specific hyperparameter tuning. Future studies may benefit from optimizing pixel-wise model configurations for each domain to maximize performance in operational settings.

Additionally, the comparison of patch-based methods is outside the scope of this study. The systematic survey and comparison of novel image patch-based methods, particularly semantic segmentation models such as U-Net or Vision Transformer (ViT), is crucial to identifying optimal practices for large-scale crop mapping. Equally important is surveying and comparing the continued advancement of transfer learning techniques, including FSL and SFUDA, which offer the potential to extend crop mapping knowledge from well-surveyed regions to broader areas, thereby supporting more inclusive crop mapping. Moreover, post-processing techniques, although not discussed, are a promising approach to improve prediction maps, especially with cadastral boundaries or high-resolution field segments. Finally, the choice of imagery product should be consistent with the structure of the landscape to reduce mixed pixels (e.g., 30-m Landsat for large fields, 10-m Sentinel-2 or 3-m PlanetScope for smallholders). High-revisit products such as Harmonised Landsat-Sentinel (HLS) provide dense spectral-temporal coverage and also have great potential for future crop mapping applications.

\section{Conclusion}

This study reviewed current practices in large-scale, pixel-wise crop mapping and transfer learning workflows, and evaluated six preprocessing methods, eleven model architectures, and three transfer learning techniques through systematic experiments. The findings offer practical guidelines for selecting optimal workflows based on the availability of labeled data and domain shifts.

Among six preprocessing approaches, 7-day linear resampling consistently delivered the most accurate and transferable results, offering a fixed temporal structure that supports generalizability across regions and years. Other fine-scale methods, such as phenological peak and smoothing, performed well in supervised settings but were less robust under diverse direct transfer. Transformer models outperformed others by effectively capturing spectral-temporal dynamics, while RF provided a competitive balance of accuracy and computational efficiency. More complex RNN variants did not yield clear performance gains, suggesting that increased model complexity does not guarantee better results. Our results showed that model performance stabilizes beyond a threshold sample size (approximately 3,000 for each class for a typical Landsat footprint in the Corn and Soybean Belt), though additional samples still improve generalizability. Complementing input with VIs and SAR data enhanced RF performance notably, but had minimal impact on Transformer models.

In transfer scenarios, direct transfers were effective only under low domain shifts. The multi-year model showed better generalizability than the single-year model in temporal transfer. UDA improved target-crop accuracy in homogeneous domains but struggled with heterogeneous classes. Fine-tuning consistently adapts pre-trained models to diverse transfer scenarios, with balanced subset undersampling enhancing robustness under highly imbalanced target-site samples. However, once the labeled sample size exceeds a certain threshold, full supervised training outperforms fine-tuning and DANN. The workflow strategy selection should remain flexible, adapting to sample availability and domain-specific conditions.

\section{Glossary}

\begin{itemize}
    \item AtBi-GRU: Attention-Bi-GRU
    \item AtBi-LSTM: Attention-Bi-LSTM
    \item AtBi-RNN: Attention-Bi-RNN
    \item Bi-GRU: bidirectional GRU
    \item Bi-LSTM: bidirectional LSTM
    \item Bi-RNN: bidirectional RNN
    \item CDL: Cropland Data Layer
    \item CNN: Convolutional Neural Network
    \item DANN: Domain-Adversarial Neural Networks
    \item DL: deep learning
    \item DT: Decision Tree
    \item DTW: Dynamic Time Warping
    \item ETM+: Enhanced Thematic Mapper Plus
    \item EVI: Enhanced Vegetation Index
    \item FSL: few-shot learning
    \item GCVI: Green Chlorophyll Vegetation Index
    \item GEE: Google Earth Engine
    \item GRD: Ground Range Detected
    \item GRU: Gated Recurrent Units
    \item HLS: Harmonised Landsat-Sentinel
    \item IW: Wide Swath
    \item LC08: Landsat 8
    \item LE07: Landsat 7
    \item LSTM: Long Short-Term Memory
    \item LSWI: Land Surface Water Index
    \item ML: machine learning
    \item MLP: Multilayer Perceptron
    \item MSI: Moisture Stress Index
    \item MTL: multi-task learning
    \item NASS: National Agricultural Statistics Service
    \item NDTI: Normalized Difference Turbidity Index
    \item NDVI: Normalized Difference Vegetation Index
    \item NDWI: Normalized Difference Water Index
    \item NIR: near-infrared
    \item OA: overall accuracy
    \item OLI: Operational Land Imager
    \item PA: producer's accuracy
    \item RF: Random Forest
    \item RNN: Recurrent Neural Network
    \item S1: Sentinel-1
    \item SAR: synthetic aperture radar
    \item SVMs: Support Vector Machines
    \item UDA: unsupervised domain adaptation
    \item ViT: Vision Transformer
    \item VIs: vegetation and water indices
    \item WE: Whittaker-Eilers
    \item 1DCNN: one-dimensional CNN architecture
\end{itemize}

\section{Acknowledgements}

This research was supported by the NASA Land Cover and Land Use Change grant \#80NSSC22K0467. We thank the BioSense Institute, University of Novi Sad for their partnership, support, and sharing of data and knowledge throughout this study. We thank Denise Heikinen for the help with English composition.

\section{Code availability}

The \href{https://github.com/JudyJuezhuLong/Best-Practices-for-Large-Scale-Pixel-Wise-Crop-Mapping-and-Transfer-Learning-Workflows.git}{GitHub repository} includes step-by-step experiment code, extended methodological details, and summarized experiment results. The full workflow and evaluation pipeline are organized into clearly structured modules corresponding to our large-scale crop classification study.

\section{Declaration of Competing Interest}

The authors declare that they have no known competing financial interests or personal relationships that could have appeared to influence the work reported in this paper.

\appendix

\section{Appendix A. Trusted pixels versus original CDL samples} \label{sec:trustedvsCDL}
\FloatBarrier

\begin{figure}
	\centering
		\includegraphics[scale=0.6]{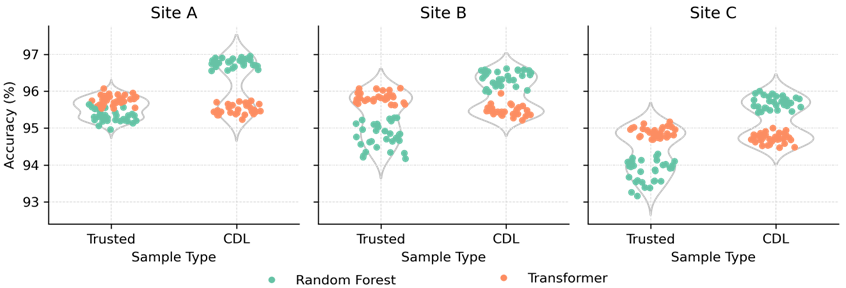}
    \caption{Overall accuracy from 10-fold cross-validation for RF and Transformer models across three preprocessing methods and two sample types at Sites A, B, and C. Each violin plot summarizes 30 accuracy values per model, comprising results from three preprocessing strategies (7-day linear resampling, WE-smoothed 7-day linear resampling, and phenological peak), each evaluated with 10 folds. Color clusters highlight model-level performance trends. Transformer models consistently achieve higher accuracy with trusted samples, while RF models perform best with CDL labels.}
	\label{fig:20}
\end{figure}

\begin{figure}
	\centering
		\includegraphics[scale=0.23]{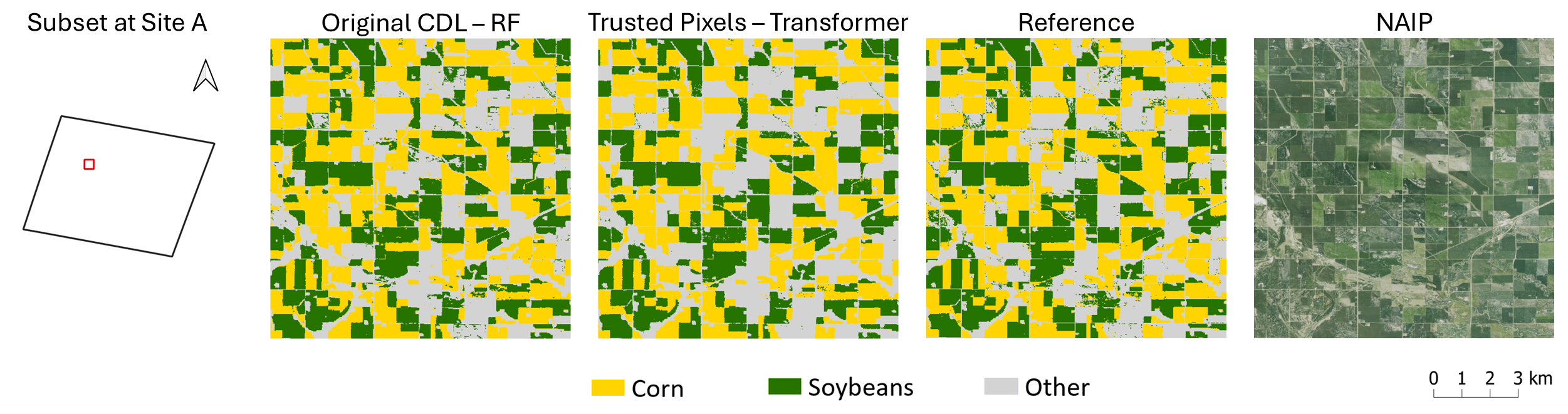}
    \caption{Comparison of model predictions using original CDL versus trusted samples for a 500-by-500-pixel subset (\textasciitilde225 km$^2$) at Site A. RF trained on original CDL and Transformer trained on trusted samples are shown alongside the 2023 CDL reference map. Transformer predictions show clearer, more coherent patterns, especially at field boundaries. Abbreviations: CDL, 2023 Cropland Data Layer; RF, Random Forest model.}
	\label{fig:21}
\end{figure}

Reliable ground reference labels are essential for accurate crop classification, but obtaining consistent and error-free labels remains challenging. The USDA CDL data, although widely used, inherently contain uncertainties, with 85\% to 95\% producer's accuracy \citep{cdl_acc}. We evaluated the "trusted pixels," which were derived using multi-year crop rotation filtering, and compared them with original (raw) CDL samples in terms of the performance of crop classification models. To perform this evaluation, we utilized a consistent experimental setup (Table~\ref{tab:5}), comparing model performances using the same total number of training samples (400,000), while the testing was uniformly carried out on the same trusted test dataset (40,000 samples) to ensure a fair comparison.

The results demonstrated clear differences in model performance between trusted pixels and raw CDL samples (Fig.~\ref{fig:20}). Specifically, the Transformer model trained on trusted pixels consistently outperformed its counterpart trained on original CDL data across all three evaluated sites (A, B, and C). This performance difference indicates that DL models, such as Transformer, strongly benefit from high-precision training samples. Conversely, the RF model yielded slightly higher accuracies when trained on original CDL data than on trusted pixels. This seemingly contradictory result can be explained by the ensemble nature of RF, which benefits from the broader spectral and temporal variability inherently present in the less-filtered, original CDL data.

However, accuracy alone does not fully capture the quality of classified maps, especially for practical agricultural monitoring. Visual inspection of crop-type prediction maps for Site A provides crucial additional insights (Fig.~\ref{fig:21}). Despite marginally higher numerical accuracy in CDL-trained RF predictions, these maps suffered from pronounced "salt-and-pepper" noise and fragmented classification boundaries. In contrast, Transformer predictions using trusted pixels exhibited clearer and more coherent spatial patterns, particularly at field boundaries, demonstrating more reliable and agriculturally meaningful delineations of major crops like corn and soybeans. Such spatial coherence and boundary accuracy are critical for real-world applications such as yield estimation.

Therefore, while original CDL samples offer a broader spectral-temporal variability beneficial to certain classifiers, the cleaner and more temporally consistent trusted pixels are ultimately superior for high-quality crop mapping, especially when spatial coherence, field boundary delineation, and practical utility are prioritized.

\section{Appendix B. Phenology analysis} \label{sec:PhenologyAnalysis}
\FloatBarrier
\begin{figure}
	\centering
		\includegraphics[scale=0.45]{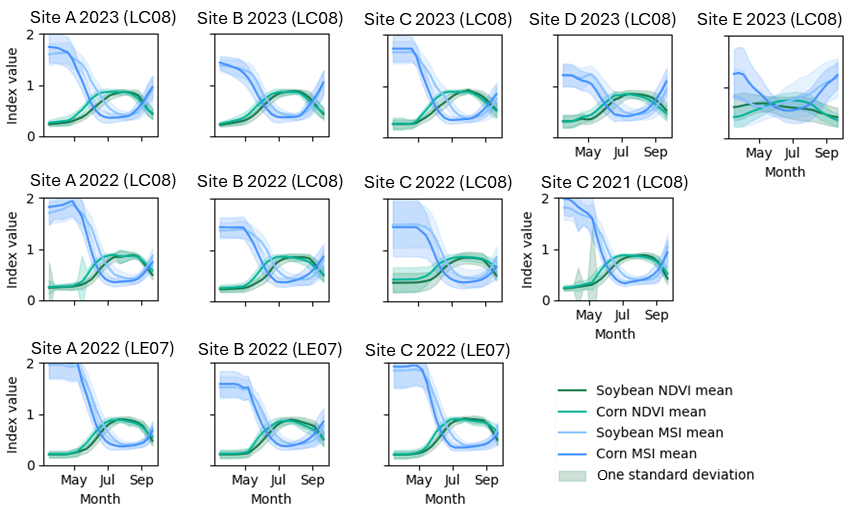}
    \caption{NDVI and MSI time series curves for corn and soybean across spatial, temporal, and sensor transfer scenarios, indicating seasonal changes in vegetation greenness and water stress. Shaded areas represent one standard deviation for all scenarios except Site E, where the shaded area represents the 95th percentile. Overall, both crops reach peak greenness (NDVI) around July, coinciding with decreasing water stress (MSI) following late-April planting. Fluctuations in magnitude and timing across sites and years reflect differences in environmental conditions. Abbreviations: NDVI, Normalized Difference Vegetation Index; higher values (green shades) indicate increased vegetation vigor and chlorophyll content; MSI, Moisture Stress Index; higher values (blue shades) indicate increased water stress.}
	\label{fig:22}
\end{figure}

\section{Appendix C. DTW analysis} \label{sec:dtwappendix}
\FloatBarrier
\begin{table}[H]
    \centering
    \caption{Mean intra- and inter-class DTW distances across six spectral bands for six preprocessing methods, calculated from randomly selected 100 corn and 100 soybean samples.}
    {\scriptsize
    \begin{threeparttable}
    \begin{tabular}{p{4cm}>{\centering\arraybackslash}p{1.5cm}>{\centering\arraybackslash}p{1.5cm}>{\centering\arraybackslash}p{1.5cm}>{\centering\arraybackslash}p{1.5cm}>{\centering\arraybackslash}p{3.5cm}}
    \toprule
    \multirow{2}{=}{\textbf{Method}} & \multirow{2}{=}{\textbf{Band}} & \multirow{2}{=}{\textbf{Intra-class (Corn)}} & \multirow{2}{=}{\textbf{Intra-class (Soy)}} & \multirow{2}{=}{\textbf{Inter-class\tnote{1}}}& \textbf{Inter-class>Intra-class Band Count\tnote{2}}\\
    \midrule
    \multirow{6}{=}{Raw}
    & Blue & 0.4306 & 0.4196 & 0.4240 &\\
    & Green & 0.3917 & 0.3867 & 0.3892 &\\
    & Red & 0.3941 & 0.3849 & 0.3916 &2\\
    & NIR & 0.3394 & 0.3553 & 0.3516 &\\
    & SWIR1 & 0.2210 & 0.2173 & 0.2304 &\\
    & SWIR2 & 0.1932 & 0.1848 & 0.2024 &\\
    \midrule
    \multirow{6}{=}{Weighted-WE smoother}
    & Blue & 0.0433 & 0.0441 & 0.0507 &\\
    & Green & 0.0478 & 0.0499 & 0.0600 &\\
    & Red & 0.0628 & 0.0711 & 0.0847 &5\\
    & NIR & 0.1481 & 0.1833 & 0.1786 &\\
    & SWIR1 & 0.1438 & 0.1661 & 0.1968 &\\
    & SWIR2 & 0.1297 & 0.1507 & 0.1794 &\\
    \midrule
    \multirow{6}{=}{7-day LN resampling}
    & Blue & 0.0598 & 0.0606 & 0.0680 &\\
    & Green & 0.0710 & 0.0754 & 0.0890 &\\
    & Red & 0.0940 & 0.1052 & 0.1220 &5\\
    & NIR & 0.1821 & 0.2362 & 0.2271 &\\
    & SWIR1 & 0.2076 & 0.2149 & 0.2529 &\\
    & SWIR2 & 0.1920 & 0.1880 & 0.2207 &\\
    \midrule
    \multirow{6}{=}{30-day LN resampling}
    & Blue & 0.0438 & 0.0393 & 0.0446 &\\
    & Green & 0.0474 & 0.0469 & 0.0518 &\\
    & Red & 0.0740 & 0.0645 & 0.0774 &5\\
    & NIR & 0.1592 & 0.1837 & 0.1764 &\\
    & SWIR1 & 0.1457 & 0.1484 & 0.1636 &\\
    & SWIR2 & 0.1534 & 0.1354 & 0.1612 &\\
    \midrule
    \multirow{6}{=}{WE smoothed 7-day LN}
    & Blue & 0.0660 & 0.0600 & 0.0661 &\\
    & Green & 0.0759 & 0.0682 & 0.0809 &\\
    & Red & 0.1023 & 0.0934 & 0.1142 &5\\
    & NIR & 0.2499 & 0.2435 & 0.2644 &\\
    & SWIR1 & 0.2318 & 0.2075 & 0.2570 &\\
    & SWIR2 & 0.2390 & 0.1856 & 0.2380 &\\
    \midrule
    \multirow{6}{=}{Phenological peak}
    & Blue & 0.0749 & 0.0853 & 0.0945 &\\
    & Green & 0.0897 & 0.0998 & 0.1134 &\\
    & Red & 0.1133 & 0.1257 & 0.1428 &6\\
    & NIR & 0.2183 & 0.2368 & 0.2676 &\\
    & SWIR1 & 0.2214 & 0.2064 & 0.2696 &\\
    & SWIR2 & 0.2125 & 0.1743 & 0.2540 &\\
    \bottomrule
    \end{tabular}%
    \begin{tablenotes}
        \item[1] Corn and soybean are the two target crops that are the most challenging to separate, so the DTW analysis only evaluates the time series of these two target crops.
        \item[2] The count of bands where the inter-class DTW is larger than two intra-class DTWs, indicating that the preprocessing method can better distinguish between corn and soybean time series.
    \end{tablenotes}
    \end{threeparttable}
    }
\label{tab:9}
\end{table}

\section{Appendix D. Overall accuracy of transfer learning approaches across spatial domains} \label{sec:transferlearningOA}

\begin{figure}
	\centering
		\includegraphics[scale=0.58]{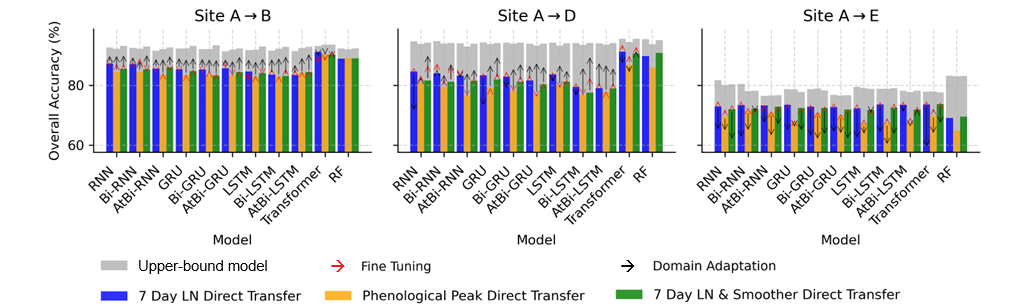}
    \caption{Comparison of transfer learning approaches across spatial domains in 2023. The grouped bar plot summarizes the overall accuracy of eleven pixel-wise classification models combined with three preprocessing methods. Grey bars indicate the test accuracy of the upper-bound model trained from scratch for each target site, while the colored bars indicate the test accuracy of the direct-transfer model. The arrow length indicates the accuracy difference between adapted and directly transferred models, highlighting the impact of transfer learning strategies. Results represent the mean overall accuracy from 8 out of 10 cross-validation folds, excluding the lowest and highest fold results to minimize the influence of extreme cases.}
	\label{fig:23}
\end{figure}

\section*{Declaration of generative AI and AI-assisted technologies in the writing process}

During the preparation of this work, the authors used ChatGPT to assist in improving the clarity and readability of the writing. The authors reviewed and edited the content and take full responsibility for the content of the publication.

\bibliographystyle{elsarticle-harv} 
\bibliography{citation_bib} 








\end{document}